\documentclass[sigconf, nonacm]{acmart}
\usepackage{tabularx}   
\usepackage{tikz}
\usetikzlibrary{positioning, arrows.meta, shapes, backgrounds}
\usepackage{graphicx} 
\usepackage{CJKutf8} 
\usepackage{subfig}
\usepackage{xspace}
\usepackage{balance}
\usepackage{stfloats}
\usepackage{amsmath,amsthm}
\usepackage[normalem]{ulem}
\usepackage{makecell}
\usepackage{array}
\usepackage{algorithm}
\usepackage{algpseudocode}
\usepackage{placeins}
\usepackage{fancybox}
\usepackage{xcolor}
\usepackage{enumitem}
\usepackage{cleveref}

\AtBeginDocument{%
  }

\usepackage{listings}
\definecolor{codecommentcolor}{RGB}{128, 67, 119}
\definecolor{codenumbercolor}{rgb}{0.5, 0.5, 0.5}
\definecolor{codestringliteralcolor}{RGB}{30, 66, 143}
\definecolor{codefieldcolor}{RGB}{34, 128, 33}
\definecolor{codepythonkeywordcolor}{RGB}{152, 59, 51}
\newcommand{\sv}[1]{\textcolor{green}{}
}
\newcommand{\aj}[1]{\textcolor{blue}{}
}
\newcommand{\tm}[1]{\textcolor{red}{}
}
\newcommand{\ach}[1]{\textcolor{purple}{}
}

\usepackage{xcolor}
\definecolor{darkpurple}{RGB}{102, 0, 153}
\newcommand{\tn}[1]{}

\newcommand{\eg}{\emph{e.g.,}\xspace}

\newcommand{\ie}{\emph{i.e.,}\xspace}

\newcommand{\toolname}{\texttt{FlexiFlow}\xspace}
\newcommand{\operator}{operator\xspace}
\newcommand{\Operator}{Operator\xspace}
\newcommand{\job}{job configuration\xspace}
\newcommand{\Job}{Job Configuration\xspace}

\newcommand{\node}{operator\xspace}

\newcommand{\snd}{{\tt SC}\xspace}

\newcommand{\mnd}{{\tt MC}\xspace}

\newcommand{\ic}{{\tt IC}\xspace}
\newcommand{\scqa}{{\tt SCQA}\xspace}
\newcommand{\lcqa}{{\tt LCQA}\xspace}
\newcommand{\icp}{{\tt ICP}\xspace}

\newcommand{\somc}{single operator, multiple choice}
\newcommand{\momc}{multiple operator, multiple choice}

\begin{document}

\title{FlexiFlow: Bandit-based Model Switching in ML Workflows}

\author{Abhilash Jindal}
\email{ajindal@cse.iitd.ac.in} 
\affiliation{%
  \institution{IIT Delhi}
  \country{India}
}

\author{Todd Nief}
\email{tnief@uchicago.edu}
\affiliation{%
  \institution{University of Chicago}
  \country{USA}
}

\author{Bhanu Prakash Vangala}
\email{bv3hz@missouri.edu}
\affiliation{%
  \institution{University of Missouri}
  \country{USA}
}

\author{Tanu Malik}
\email{tanu@missouri.edu}
\affiliation{%
  \institution{University of Missouri}
  \country{USA}
}

\author{Shankaradithyaa V}
\email{shankaradith@iisc.ac.in}
\affiliation{%
  \institution{IISc Bangalore}
  \country{India}
}

\author{Tvisha Malik}
\email{cs1210085@cse.iitd.ac.in}
\affiliation{%
  \institution{IIT Delhi}
  \country{India}
}

\author{Anshik Sahu}
\email{cs1210577@cse.iitd.ac.in}
\affiliation{%
  \institution{IIT Delhi}
  \country{India}
}

\author{Amitabh Chaudhary}
\email{amitabh@uchicago.edu}
\affiliation{%
  \institution{University of Chicago}
  \country{USA}
}

\author{Aaron Schein}
\email{schein@uchicago.edu}
\affiliation{%
  \institution{University of Chicago}
  \country{USA}
}

\begin{abstract}
Model optimizations help improve inference performance and accuracy of ML workflows. 
However, relying on a single model to perform inference across all data batches
often fails to maximize accuracy and thus overall performance. In many cases,
alternate models could perform better on specific subsets of data where a
primary model underperforms. Our experiments with real ML workflows indeed show that
switching models  improves workflow accuracy by up to 23\%. 

Yet, current systems lack the ability to adaptively switch between models based
on performance, forcing users to manually test models in sequence. We present
\toolname, a dataflow system that dynamically switches between alternate models
when the current model exhibits low accuracy. 
\toolname learns to rank models using a novel multi-armed bandit approach that
accounts for model runtimes, probability of passing user-defined assertions, 
and the computational structure of the ML workflow. 
We show that the standard Thompson sampling approach is insufficient for
switching models in ML workflows. In contrast, our proposed approaches are
effective and scales to complex real-world ML workflows. Experiments show that
switching models at runtime while reusing intermediate results provides higher accuracy, but also 
48\% efficiency gain compared to sequential workflow runs. 



\end{abstract}

\renewcommand\footnotetextcopyrightpermission[1]{} 
\settopmatter{printacmref=false} 
\settopmatter{printfolios=true} 
\maketitle

\section{Introduction}
\label{sec:intro}

With the increasing use of machine learning in production systems, engineers often encounter practical problems in deploying and maintaining ML models and workflows. Traditionally, models are often learned in notebooks or scripts, serialized (i.e., “pickled”), and embedded
within web servers---\textit{a.k.a} deployed---to expose prediction APIs. However, deployment often  reveals a range of issues, particularly related to model performance and accuracy. 

A recurring concern is the issue of soft failures---situations where the model does not crash but delivers incorrect or degraded predictions, often due to factors like data drift~\cite{gao2008classifying}. These soft failures compromise prediction accuracy and, in turn, the overall effectiveness of the deployed system. When workflows consist of multiple steps and models, such degradations must be promptly identified and addressed. Developers commonly rely on trace logs to identify problematic data batches, but exhaustively testing alternative model configurations is resource-intensive and time-consuming. 

\begin{figure}
    \setlength\belowcaptionskip{-15pt}
    \centering \includegraphics[width=0.9\linewidth]{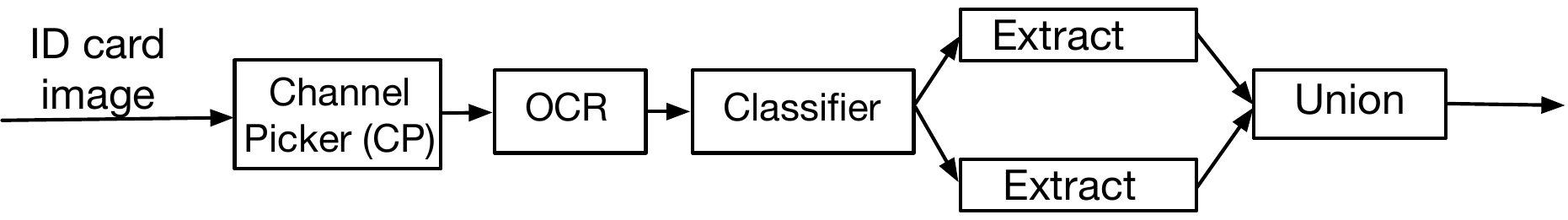}
    \caption{ID Card ML workflow. 
    }
    \label{fig:IDworkflow}
\end{figure}

Consider a deployed ML workflow, visualized in Figure~\ref{fig:IDworkflow}, for identity verification in KYC (Know Your Customer) scenarios using identification cards (e.g., driver’s licenses, passports). The workflow processes an image or scan of the ID card and outputs a structured record with fields such as name, date of birth, ID number, and expiration date. Each step in this pipeline—such as image cropping, color channel selection, OCR, card classification, filtering, and field extraction—may be implemented using different ML models. For instance, the cropping step could use models like YOLO~\cite{redmon2016you}, SSD~\cite{liu2016ssd}, or Faster R-CNN~\cite{ren2015faster}, while OCR could be performed by tools such as Tesseract, EasyOCR, Google Vision API, AWS Textract, or Microsoft Read API. In practice, only one \emph{single} model is typically selected for each step 
based on evaluation over training data. However, this static selection often fails to maintain high accuracy across all real-world inputs during deployment. If a single model underperforms, the entire workflow may need to be re-executed from the beginning with an alternative model, significantly increasing the management burden and latency.


One approach to mitigating soft model failures on specific inputs is to dynamically switch between models based on their performance. For example, in the OCR step of an ID card processing pipeline, if EasyOCR fails on a given input, the system can switch to an alternative like Tesseract. However, this requires the system to assess relative model performance—not just using aggregate accuracy metrics, but by detecting soft failures (e.g., invalid outputs or low confidence) on individual inputs. Even when estimates of model performance are known, a key challenge remains: determining the optimal order in which to try models to minimize latency while maintaining correctness.


Recently, these challenges have been addressed, albeit  independently, in the machine learning domain. Model assertions~\cite{kang2020model} provide a data-independent way to assess model performance. In the context of ID card processing, such assertions might include checks like validating that the extracted date of birth is a legitimate calendar date, ensuring the age is under 120 years, or applying confidence thresholds to predictions. Separately, explore-exploit strategies~\cite{Sutton1998} offer a principled method for 
determining which configurations to use first. These approaches balance exploiting configurations that have performed well historically with exploring new alternative configurations, adapting dynamically based on observed rewards.
To our knowledge, no existing ML dataflow system integrates both model assertions and explore-exploit strategies to guide model switching decisions. Integrating model assertions with explore-exploit strategies requires a non-trivial coordination of correctness and efficiency, as the system must adapt to assertion violations while maintaining acceptable performance. 

In this paper, we present \toolname 
a dataflow system for multi-step ML workflows that learns to 
perform model selection and switching when models (soft) fail on assertions.  \toolname allows a user to configure each step of an ML workflow with multiple model choices. To determine the best model configuration, \textit{i.e.,} combination of model choices at each step, we equip \toolname with a Bayesian statistics-based optimizer
that samples from the posterior reward distributions of model choices, enabling the estimation of the most effective model configuration for a given input. If a model configuration still fails assertions, to explore alternative configurations, \toolname rolls back to a previous model operator and efficiently tests a sequence of new
configurations.  We demonstrate that a variety of explore-exploit Bayesian strategies 
are possible within the runtime to enhance performance.
Our findings indicate that while the space of all possible workflow
configurations can be large, Bayesian strategies can quickly find good
configurations.  

\subsection{Our Contributions}

Figure~\ref{fig:FF-IDcard} shows the different choices and assertions specified
for an instance of the ID card ML workflow. In the ChannelPicker step, five color channel
choices are made available: all channels (all), blue only (b), green only (g),
red only (r), and a combination of blue and green channels (bg).
In the OCR step, the user provides choice between two engines:
Tesseract and EasyOCR.  Additionally, the workflow allows for different
classifier choices in the classifier step. The figure also depicts assertion
checks on extracted data. As shown, there are a total of 20 different workflow
configurations, corresponding to 5, 2, and 2 operator choices at various steps.
The configuration space grows quickly as the number of choices increases or as
more steps are added. 


Although an ML workflow might select models or configurations based on performance during training, these choices often degrade when deployed, due to shifts in the data distribution. To remain effective, the system must adapt to changing conditions at inference time. This motivates the use of online learning approaches where, rather than training the model selection process all at once on a fixed dataset, the system incrementally updates its model selection process with the arrival of each new example or batch of examples.

A common approach to online learning is the multi-armed bandit (MAB), where the goal is to choose among $k$ configurations by balancing exploration and exploitation to maximize expected reward. However, \toolname{} departs from the standard MAB setting in two key ways:

\vspace{3pt}
\noindent First, while there are $k$ overall workflow configurations, each configuration is composed of multiple steps (or operators), and each step has its own set of model choices. This implies that a distinct MAB problem exists at each stage of the workflow. This MAB problem likely includes interactions across the decision at each step, resulting in a bandit problem that does not fit neatly into the standard contextual or hierarchical bandit structure.

\vspace{3pt}
\noindent Second, unlike traditional MAB settings where each trial is independent, in \toolname, if a given configuration fails (e.g., violates a model assertion), it should not require restarting the entire workflow. Instead, the system should be able to `rollback’ to the point of failure and selectively modify only the relevant operator choice. For instance, as shown in Figure~\ref{fig:FF-IDcard}, if a classification step yields incorrect results, the system should be able to switch classifiers without re-running upstream components like OCR or preprocessing. 

These distinctions highlight the need for a bandit-based approach that leverages the computational structure of the workflow graph, enabling localized reconfiguration and efficient exploration of the model choice space for improved overall performance. We make the following contributions:

\begin{figure}
    \setlength\belowcaptionskip{-15pt}
    \centering
    \includegraphics[width=\linewidth]{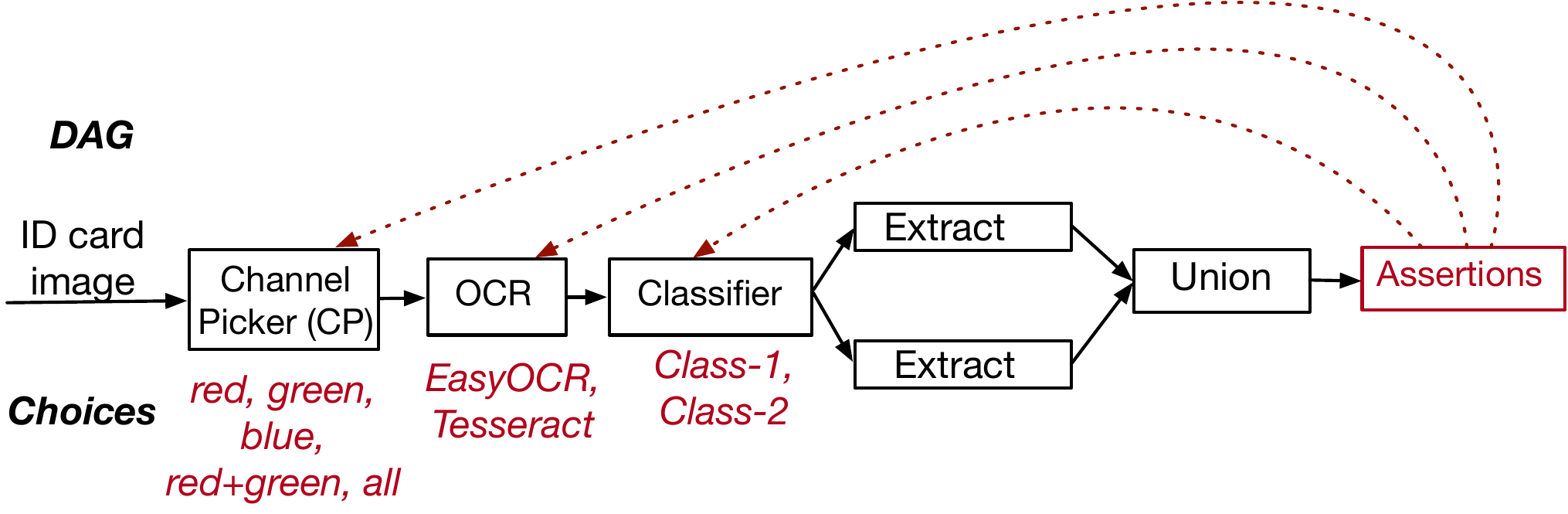}
    \caption{ID Card workflow in \toolname. Choices for each workflow step are indicated below in red italics. Soft failure of assertions necessitates switching (dashed arrows) between model choices.}
    \label{fig:FF-IDcard}
\end{figure}

\begin{itemize}[leftmargin=*]

\item {\bf A novel programming model with operator choices and assertions:}
We introduce primitives within an existing dataflow language to support the expression of both operator
choices and assertions. 
The language runtime, to support multiple choices, incorporates an adaptive
rollback mechanism 
that reverts to specific  operators when an assertion
fails. This rollback mechanism is guided by an optimizer (described next), allowing for rollbacks
to intermediate nodes within the workflow and not just the root step. To
facilitate incremental computation after rollback, we leverage past work on dataflow error handling~\cite{deem24reactive}. 


\vspace{5pt}
\item {\bf Bayesian strategies for automatic error correction:} We
propose two multi-armed bandit-based strategies to learn an optimal sequence of
configurations:

\begin{itemize}

\item The first configuration strategy uses Thompson sampling with posterior distributions
over model configurations, balancing exploitation of high-payoff configurations that
pass the most assertions with exploration of alternative configurations. Execution time is
explicitly modeled as part of the optimization strategy.

\item The second hybrid strategy exploits the structure of the choice-based programming model 
by running separate multi-armed bandit  optimizations at each workflow step, tracking
cumulative payoffs for configurations and updating posteriors for operator
choices on high-payoff model configurations. 

\end{itemize}

Using microbenchmarks and real applications, we show that the hybrid (second)
strategy provides a learning advantage.

\vspace{5pt}
\item {\bf A research prototype:} We present \toolname, an
experimental Python prototype that supports efficient rollbacks, incremental
computation, and the proposed language extensions and optimization strategies.
We evaluate \toolname with three types of ML workflows: document processing,
image classification, and question-answering. For each workflow, we describe its
formulation in \toolname, the range of choices and assertions, and the
performance of our proposed strategies. Our key findings are:
\begin{itemize}
    \item Adding operator choices in workflows can improve $F_1$ scores by upto
    23\% compared to workflows without them. We find that this is because
    different configurations, with different ML models, often have uncorrelated
    failures. 
    \item It is imperative for strategies to learn runtimes of operator choices
    along with their probabilities of passing assertions. Further, by making
    them aware of the rollback approach of \toolname, they can quickly learn
    good configurations.
\end{itemize}
\end{itemize}

The rest of the paper is organized as follows: Section~\ref{sec:FlexiFlow} presents the
\toolname\ architecture and describes the programming model consisting of 
operator choices, assertions, and rollbacks. Section~\ref{sec:problem} formulates the
configuration selection learning problem in terms of assertions and rollbacks.
In Section~\ref{sec:strategies} we describe Bayesian strategies that estimate the probability
of choosing a configuration with the highest reward.  In Section~\ref{sec:experiments} we
present our  implementation of \toolname\ and evaluation on microbenchmarks and four
real ML pipelines.  We discuss our results in Section~\ref{sec:discussion}, present related
work in Section~\ref{sec:relatedwork}, and conclude in Section~\ref{sec:conclusion}.

\section{The \toolname Execution Model}
\label{sec:FlexiFlow}

\toolname is defined within the context of a dataflow language that enables users to express complex machine learning (ML) workflows. We begin by reviewing the fundamentals of such a language and its data model, and then introduce two novel programmatic constructs provided by \toolname: operator choices and workflow assertions.

A basic dataflow language, such as Spark-SQL~\cite{armbrust2015sparksql} or SCOPE~\cite{zhou2012scope}, models an ML workflow as a directed acyclic graph (DAG). In this DAG, nodes represent operators, each of which performs a data transformation; edges represent data dependencies between operators (i.e., the flow of intermediate datasets). Let $d$ denote an input data item (e.g., a row or an image), and let $op$ denote a map transformation function implemented by an operator. Then, the semantics of an operator are defined as: 

\begin{center}
$op(d) \rightarrow d’$ 
\end{center}

\noindent in which $d'$ is the output data item produced by applying $op$ to $d$. The execution of a dataflow program thus proceeds by propagating data from source nodes to sink nodes, applying each operator as data becomes available. The runtime supports pipeline parallelism~\cite{hennessy2017computer}: partial outputs from parent operators can be streamed to child operators without waiting for the entire input to be processed.

\begin{figure}[t]
    \setlength\abovecaptionskip{-1pt}
    \setlength\belowcaptionskip{-12pt}
    \centering
    \includegraphics[width=\linewidth, page=1]{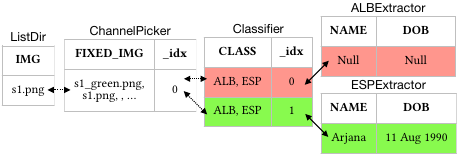}
    \caption{\toolname data model. {\small Some operators are omitted for clarity.}}
    \label{fig:ffdatamodel}
\end{figure}

Typical operators are data preprocessing operators such as extracting structured rows from a file, processing them, combining rows, or outputting to disk or user-defined operators such as processing video frames or images into structured data. 
Each operator consumes a well-defined input schema and produces a corresponding output schema as shown in Figure~\ref{fig:ffdatamodel}. The runtime verifies schema compatibility along edges in the DAG to ensure compositional correctness. In particular, operators are map operators that  apply the operator independently to each data item, and currently do not include logical operators such as joins, and aggregates. 

\toolname augments this basic model by introducing three core constructs: operator choices and workflow assertions as language constructs, and rollback as a runtime construct. In \toolname, each map operator $op$ may be instantiated by one of multiple alternative implementations 
Let $\{op^1, op^2, \ldots, op^k \}$ 
denote the $k$ possible implementations. The operator is now modeled as a non-deterministic transformation, selected at runtime:

\begin{center}
$op^i (d) \rightarrow d’^i$, where $i \in {1,…,k}$ 
\end{center}

\noindent That is, for a given input $d$, the runtime selects an instance choice $op^i$ and produces the corresponding output $d'^i$. Figure~\ref{fig:FF-IDcard} also shows (in italics) multiple  choices available for each operator. 

In addition to operator choices, \toolname allows users to define assertions on intermediate or final outputs of the workflow. These are Boolean-valued functions $\phi(d') \in \{ \text{true}, \text{false} \}$ that act as semantic checks (e.g., validating that a date of birth is valid or that confidence scores exceed a threshold). 
In general there can be many assertions at different locations of a DAG, but for ease of exposition we assume a single assertion represented as a sink node in a DAG.  Assertions provide a 
signal of correctness and are used by the runtime to detect soft failures and guide, as we describe in the next section,  operator re-selection.  

Finally, when an assertion fails for a given input, \toolname's runtime rolls back to an operator. Figure~\ref{fig:FF-IDcard} shows that ID card can possibly rollback to Channel Picker, OCR or Classifier operators. We describe further implementation details of how to rollback in Section~\ref{sec:experiments}.  

\begin{figure}[t]
  \setlength\belowcaptionskip{-12pt}
    \centering
 \includegraphics[width=0.65\linewidth]{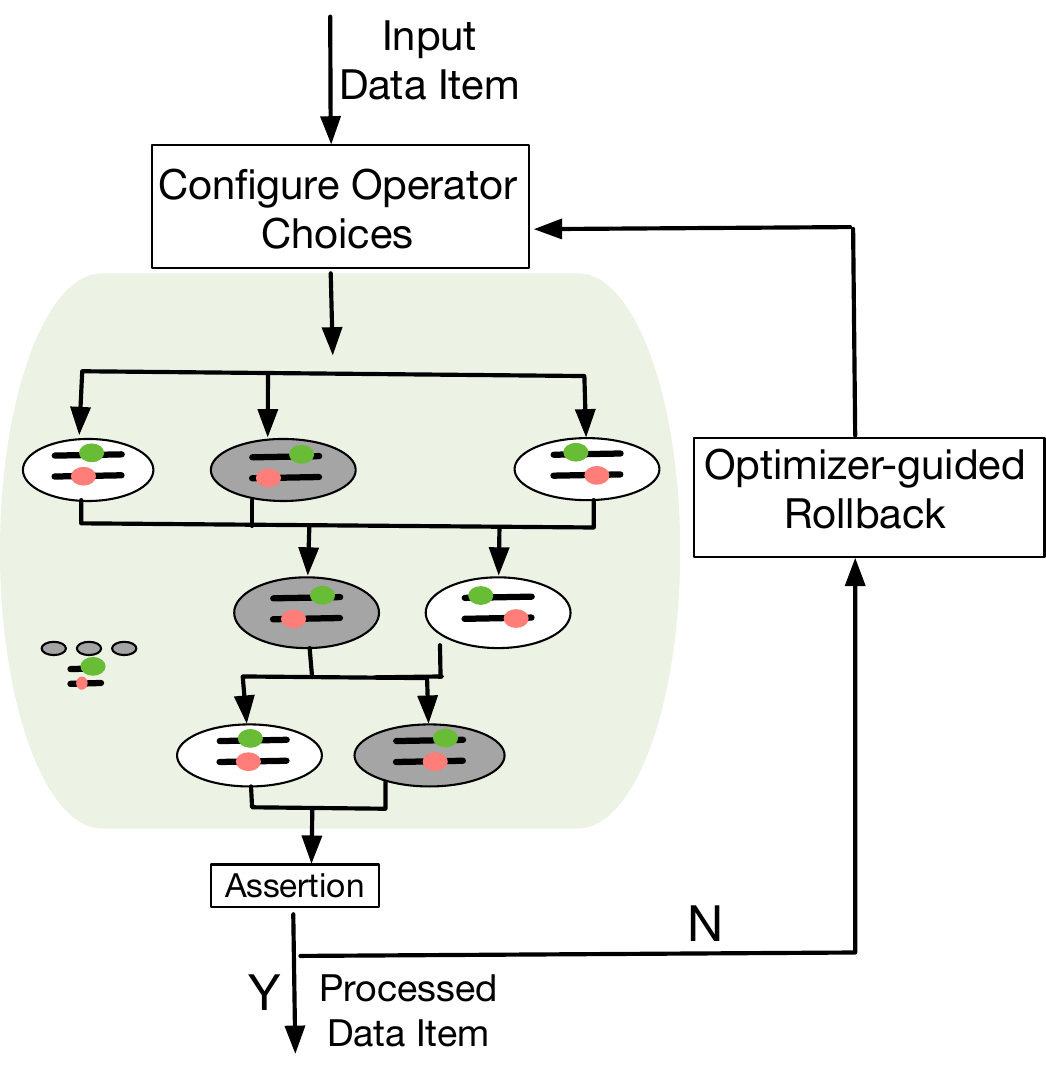}
    \caption{\toolname\ Architecture. Optimizer maintains estimates for each operator choice and set of operator choices.}
    \label{fig:ffarch}
\end{figure}

\noindent {\bf Architecture.} Figure~\ref{fig:ffarch} shows the overall architecture of \toolname with the three constructs: 
operator choices, assertions, and rollback, which introduce in \toolname the decision of which specific operator to rollback to and which operator choice to pick next, when an assertion fails. To address this, \toolname employs a runtime-aware, rollback-aware, learning-based optimizer that learns from past successes and failures at both the operator level and across combinations of operator choices as described next.  


\section{The \toolname Learning Problem}
\label{sec:problem}

The choices available at each operator in \toolname\ multiplicatively increase the space of ML workflow \textit{configurations}. 

Let an ML workflow consist of $M$ operators, where each
operator $i$, with $i \in \{1, \ldots, M\}$,  has $k_i$ possible choices. Then, the total number of workflow configurations is given by:
\begin{equation}
K = \prod_{i=1}^{M} k_i
\end{equation}
For example, the ID card processing workflow has $M = 3$ operators--- {ChannelPicker, OCR, classifier}---each having 5, 2, and 2 choices, respectively. One possible
configuration within the workflow is [``red", ``Tesseract", ``class 1"]. In total, the workflow has $5\times 2\times
2 = 20$ configurations to choose from.

For a given data item, \toolname initially selects a  configuration from the set of $K$ configurations, with the hope that the selected configuration  passes the assertion. 
If this selection fails the assertion, the system selects another configuration, 
continuing until the assertion is satisfied. This process forms a sequence of configurations:
$S = (a_1, \ldots, a_j), j \leq K$. 
In the worst case, the sequence can be as long as $K$, meaning all configurations of an ML workflow must be tested. However, the objective of \toolname is to determine the optimal sequence $S^*$  that 
minimizes the expected execution cost, as each additional configuration incurs an overhead of re-running the operators. 

We observe two properties of any sequence $S$: 
\begin{itemize}[leftmargin=*,noitemsep]
\item First, in any sequence of length $j$, the current configuration $a_j$ is tried if all previous configurations from index 1 to $j-1$   have failed the assertion. 

\item Second, 
if two consecutive configurations share a longer prefix of common operator choices, the rollback execution is faster. 
Consider the sequence
 ([``red'', ``EasyOCR'',
``class 2''], [``red'', ``EasyOCR'', ``class 1'']) and the sequence ([``red'', ``Tesseract”,
``class 2''], [``red'', ``EasyOCR'', ``class 1'']). 
The first sequence, due to longer common prefix, leads to a lower re-execution cost, improving overall efficiency. 
\end{itemize}


Given these properties, the goal in \toolname is to identify the best configuration sequence
 $S^{*}$ 
that minimizes the expected runtime, formulated as:
\begin{equation}
\label{eq:sequence_runtime}
\sum_{j \in \{1..K\}}{P \left [\bigcap_{i < j} \overline{a_i} \right ] \times  \mathcal{T}_{a_{j-1} \rightarrow {a_j}}},
\end{equation}
in which, $P \left [\bigcap_{i < j} \overline{a_i} \right ]$ is the probability that all previous configurations $\{a_1, \ldots, a_{j-1} \}$ fail the assertion, and $\mathcal{T}_{a_{j-1} \rightarrow {a_j}}$ is rollback execution cost from configuration $a_{j-1}$ to $a_j$.

Expression \ref{eq:sequence_runtime} is of limited practical use, since failure probabilities are  unknown for new data items, and expected runtimes in deployment often deviate from training phase estimates due to differences in data distributions. 
So we next describe how \toolname\ efficiently identifies $S^{*}$ from a large space of configuration sequences using an online, Bayesian explore-exploit learning strategy.

\section{\toolname's Bayesian Online Optimizer }
\label{sec:strategies}

In this section, we develop a variety of optimizers, each of which
uses a Bayesian explore-exploit learning strategy from the 
multi-armed bandit framework~\cite{Sutton1998}. 
The optimizers estimate success probabilities and runtimes to minimize total
expected runtimes (Equation~\ref{eq:2}). Below, we provide a
brief background on the Bernoulli bandit problem, followed by \toolname online
optimizers. 

\vspace{5pt}
\noindent {\bf Background.} Consider a scenario where an operator has $K$ possible choices. Each action $a \in \{1, \ldots, K\}$ has an unknown success probability $\theta_k$ meaning that each time action
$a$ is chosen, it yields a binary reward $r$. $r$ is 1 with probability $\theta_a$ (success) and $r$ is 0 with probability $1- \theta_a$ (failure). The objective of the bandit framework is to learn the success probabilities $(\theta_1, \ldots, \theta_K)$
through experimentation, while maximizing cumulative rewards over $N$ trials, where $N \gg K$.

\begin{figure}[t] 
    \setlength\belowcaptionskip{-12pt}
    \centering 
    \includegraphics[width=0.8\linewidth]{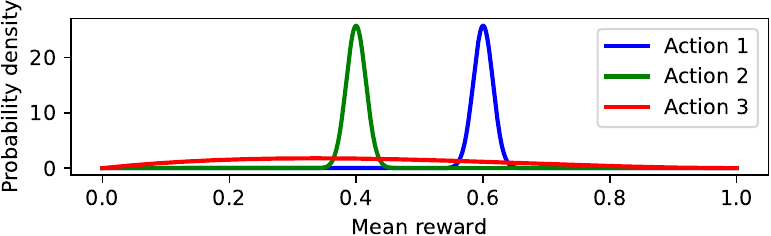} 
    \vspace{-0.1in}
    \caption{Probability density functions over mean rewards} 
    \label{fig:TS-sim} 
\end{figure}

Suppose we have an operator with three possible actions. Lets say over 2005 trials, these actions were selected as follows: Action 1 was chosen 1000 times, yielding 600 successes (${\theta}_1 = 0.6$), Action 2 was chosen 1000 times, yielding 400 successes (${\theta}_2 = 0.4$) , and Action 3 was chosen only 5 times, yielding 2 successes (${\theta}_3 = 0.4$, but with high variance due to the small sample size).
Figure~\ref{fig:TS-sim} illustrates the probability density functions (PDFs) for these estimated success probabilities. The variance in the estimates decreases as the number of trials per action increases. 


In an $\epsilon$-greedy approach, the learning agent explores by sampling actions uniformly at random with probability $\epsilon$, where $\epsilon$ is fixed a priori. During exploitation (with probability $1 - \epsilon$), the agent repeatedly selects the action with the highest estimated mean reward. For example, if action 1 consistently has a higher mean than actions 2 and 3, it will be chosen repeatedly during exploitation, while during exploration all three actions are sampled uniformly.

Allocating an equal number of experimental trials to each action during exploration results in poor performance of an  $\epsilon$-greedy approach~\cite{russo2020tutorialthompsonsampling}. Thompson Sampling (TS) fundamentally addresses this limitation of $\epsilon$-greedy by updating the probabilities of success of actions using Bayes' rule.  In Thompson Sampling, each action $a$ has an independent prior belief over each $\theta_a$, which is modeled as a Beta distribution: 
\begin{equation}
f(\theta_a; \alpha_a, \beta_a) = \frac{1}{B(\alpha_a, \beta_a)} \: \theta_k^{\alpha_a -1}\ (1-\theta_a)^{\beta_a -1}
\end{equation}
where the beta function $B(\alpha_a, \beta_a)$ is a normalizing constant. 
As observations are gathered, each action’s posterior distribution is updated according to a simple rule: 
 \begin{equation}
 (\alpha_a, \beta_a) \leftarrow 
 \begin{cases}
   (\alpha_a, \beta_a), & if\ x_t \neq a \\ 
   (\alpha_a, \beta_a) + (r_t, 1- r_t), & if\ x_t = a
 \end{cases}
 \end{equation}
 
\noindent in which $r_t$ represents the actual reward in the trial $t^{th}$ that attempts action $x_t$. 
When $\alpha_k = \beta_k = 1$, the prior probability of $\theta_k$ is uniform over [0, 1]. For each data item, only the Beta parameters ($\alpha_a$, $\beta_a$) of the selected action $a$ are updated based on the observed reward $r_t$. If the action succeeds ($r_t$ = 1), $\alpha_k$ increases, otherwise $\beta_k$ increases.  As the number of trials grows, the distribution mean of the Beta distribution $\frac{\alpha_k}{(\alpha_k + \beta_k)}$, which represents the ratio of success to total trials, approaches the empirical mean of successes. Instead of explicitly alternating between exploration and exploitation phases (as in $\epsilon$-greedy), Thompson Sampling samples from the posterior distribution of each action’s success probability $\theta_a$. Actions with higher success probability estimates are sampled 
more frequently, but uncertainty in estimates naturally encourages exploration.

\subsection{\toolname Optimizer for Single Operator with Multiple Choices}
Traditional Thompson Sampling (TS) updates success probabilities based on the observed reward from a single action/operator choice at each trial, accumulating cumulative rewards over time. However, \toolname introduces a key difference: when an operator choice fails, alternative choices are repeatedly attempted via rollbacks until a successful configuration is found. These rollbacks introduce additional runtime costs, as each re-execution incurs the overhead of rerunning the operator choice.
To address this, we first describe a modified version of Thomson Sampling (TS), termed Operator-level TS (OTS), in which only the reward function is changed and is modeled with expected runtimes of probability of success or failure of the assertions. We consider OTS for a single-operator scenario with multiple choices. We then extend this algorithm to include the computational structure of the workflow in operator and choices along with rollbacks in a multiple operator, multi-choice scenario. 



\vspace{5pt}
\noindent {\bf  Operator-Level TS (\textbf{OTS})}
In Listing~\ref{alg:somc}, each operator choice $a \in A$ has an \textit{unknown probability of passing} $ \theta_a$, modeled as a beta distribution, and an \textit{unknown runtime distribution}, modeled as a log-normal distribution. 
The domains of each distribution: beta distribution has  0-1 and log-normal has $0-\inf$, make them suitable for modeling successes and runtimes~\cite{Sutton1998}.
 To estimate these parameters before selection, the following steps are performed: We first sample the probability of passing $\theta_a$ from a Beta distribution (Line~\ref{line:sample_theta}). We then compute the expected number of runs $n_a$ to pass an assertion, which is the inverse of the probability of success (Line~\ref{line:calculate_na}). We then sample operator execution runtimes from a log-normal distribution $n_a$ times, and compute a normalized expected runtime to pass the assertion (Line~\ref{line:sample_runtime}). 
 Finally, the total estimated runtime for operator choice $a$ is given in Line~\ref{line:mean_runtime}, and for each $a$ the runtimes  is stored in $\boldsymbol{\widehat{\mathcal{T}}}$. 

Once the estimated runtimes for all available operator choices are computed, the algorithm iteratively selects, executes, and updates operator choices until the example passes. The operator with the lowest estimated runtime is chosen as $a_t^*$ in Line~\ref{line:select_action}. The data example $x_i$ is processed using $a_t^*$, and its reward (success/failure) and the runtime is observed (Line~\ref{line:observe_outcome}).  

After each operator execution, the algorithm updates its internal estimates: it refines the success probability using a Beta distribution update (Line~\ref{line:update_beta}) and adjusts the log-normal runtime distribution parameters $(\mu_a, \sigma_a^2)$ based on the observed runtime $y_t$ using the Bayesian update rule~\cite{degroot2011probability} in UpdateLogNormalParams  (Line~\ref{line:update_ln}).  
If an operator choice fails, the corresponding action is removed from consideration, and the algorithm selects another alternative (Line~\ref{line:remove_action}) \textit{i.e.,} it incorporates rollback. This process repeats until the example successfully passes the assertion.

\begin{algorithm}
\caption{Single Operator-based Thompson Sampling ($X$, $A$)}
\label{alg:somc}
\begin{algorithmic}[1]
\State \# Iterate over $N$ examples
\For{$x_i \in X$}
    \State $A_i \gets A$ \Comment{{\small Initialize action list for current example}}
    \For{$a \in A_i$}
        \State Sample $\theta_a \sim \text{Beta} (\alpha_a, \beta_a)$ \Comment{{\small Sample  probability of passing}}\label{line:sample_theta}
        \State $n_a \gets \lceil 1 / \theta_a \rceil$ \Comment{{\small Compute expected number of runs to pass}}\label{line:calculate_na}
        
        \State Sample $\tau_j \sim \text{Log-normal}(\mu_a, \sigma_a^2)$ for $j = 1, \dots, n_a$ \Comment{{\small Sample runtime for each expected run}}\label{line:sample_runtime}
        
        \State $\mathcal{T}_a \gets \frac{1}{n_a} \sum_{j=1}^{n_a} \tau_j$ \Comment{{\small Compute mean sampled runtime}}
        \label{line:mean_runtime}
        
        \State $\widehat{{\mathcal{T}_a}} \gets \mathcal{T}_a \cdot \frac{1}{\theta_a}$ \Comment{{\small Calculate estimated runtime}}\label{line:normalize_runtime}
        \State Store $\widehat{{\mathcal{T}}_a}$ in $\boldsymbol{\widehat{\mathcal{T}}}$
    \EndFor
    \State
    \While{example not passed}
        \State $a_t^* \gets \underset{a \in A_i}
        {\textrm{argmin$_a$}} \: \boldsymbol{\widehat{\mathcal{T}}}$ \Comment{{\small Select best available action (fastest runtime); Subscript $t$ indexes the processing attempt }}\label{line:select_action}
        \State
        \State Process $x_i$ using $a_t^*$; observe pass/fail $r_t$ and runtime $y_t$\label{line:observe_outcome}
        \State
        \State $(\alpha_{a_t^*}, \beta_{a_t^*}) \gets (\alpha_{a_t^*} + r_t, \beta_{a_t^*} + 1 - r_t)$ \Comment{{\small  Update Beta distribution for $a_t^*$}}\label{line:update_beta}
        \State
        \State $\text{UpdateLogNormalParams}(a_t^*, y_t)$ \label{line:update_ln}
        \State 
        \State $A_i \gets A_i \setminus \{a_t^*\}$\Comment{\small{Remove chosen action from available actions}} \label{line:remove_action}
    \EndWhile{}
\EndFor
\end{algorithmic}
\end{algorithm}

\subsection{\toolname Optimizers for Multiple Operators with Multiple Choices}
\label{sec:momc}

Thus far, we have focused on the case of a single \node{}, but in practice, workflows consist of multiple operators, each offering several choices.

One approach to learning in the \momc{} scenario is to assume operator-level independence and select the best choice for each operator individually. If the resulting configuration fails the assertion, the system proceeds to the next best configuration.

For example, in case of ID card processing, we have three operators $A = Channel\ Picker$, $B = OCR$ and $C = Classifier$ each with 5, 2, and 2 choices. To start, we determine the best operator choice (lowest expected runtime) independently at each of the three operators. Let’s say configuration $K_1 = [A_3, B_2, C_2]$ where $A_3$, $B_2$, $C_2$, represent choices within their respective operators, is the best operator choice configuration i.e., resulting in the lowest expected runtime of 10s. However, this configuration also fails the assertion. If $K_2 = [A_3,B_2,C_1]$ and $K_3 = [A_1,B_2,C_1]$ are other available configuration choices that result in 11s and 10.5s, then per OTS in Listing~\ref{alg:somc} we will choose configuration $K_3$, which has the next lowest runtime. 

Naively applying OTS at each operator independently has two drawbacks. First, assuming operator level independence may not be true.  For example, in ID card processing, a specific channel color may be necessary for EasyOCR engine to successfully extract text but that specific channel always causes Tesseract to fail. Second, choosing each operator independently is not  rollback-cost-aware, i.e. in our example, minimizing overall configuration level runtime results in choosing $K_3$ as the next best configuration even though the cost of incrementally running $K_2$ may be the lowest.  Note that $K_1$ and $K_2$ share an operator prefix and the rollback cost from $K_1$ to $K_2$ maybe lower than overall cost of incurring 10.5s for running the full configuration. 

We fix these two drawbacks next. \\
\noindent {\bf Configuration-level  TS (\textbf{CTS}).} 
One option is to relax operator level independence and learn at the
entire configuration-level. Additionally, failed configurations perform a
conditional rollback, i.e., rollback to a configuration which is estimated to
have the smallest increase in runtime from the current configuration.
Listing~\ref{alg:momc} describes learning at the configuration level using
\toolname-based TS where the set of all actions $A$ are the set of all
$K$ configurations. For each configuration, we maintain a Beta distribution with
parameters $\alpha$ and $\beta$ and obtain the estimated runtime of a
configuration $\boldsymbol{\widehat{\mathcal{T}}}$. Each configuration is sampled in
\texttt{SelectAction} to obtain a configuration with the lowest runtime,
conditional on the current best choice. The example is processed with this
configuration and actual rewards and operator-level runtimes are observed. The
Beta distribution parameters are updated similarly to Listing~\ref{alg:somc}
Line~\ref{line:update_beta}, but runtime $\mu$ and $\sigma$ are updated for each
operator choice in the current best configuration. If the example fails to pass
assertions, the next best configuration with the lowest incremental increase in
runtime is chosen in \texttt{SelectAction}. This is because our objective in Equation~\ref{eq:sequence_runtime} is to minimize the cost of rollbacks, $\mathcal{T}_{a_{t-1} \rightarrow a_{t}}/\theta_t$.

\begin{algorithm}
\caption{\Job{} Selection(X, A)}
\label{alg:momc}
\begin{algorithmic}
    \For{$x_i \in X$}
        \State \# Estimate rewards and runtimes for all actions in A
        \State $\boldsymbol{\hat{\theta}} \gets \text{EstimatePassProb}(A)$
        \State $\boldsymbol{\widehat{\mathcal{T}}} \gets \text{EstimateRuntimes}(A, \boldsymbol{\hat{\theta}})$
        \While{example not passed}
            \State \# Note: $a_t^*$ is conditional on the rollback cost from $a_{t-1}^*$
            \State $a_t^* \gets \text{SelectAction}(\boldsymbol{\widehat{\mathcal{T}}}, a_{t-1}^*)$\Comment{ $a_t^*$ is the chosen action}
            \State
            \State Process $x_i$ using $a_t^*$; observe pass/fail $r_t$ and runtime $y_t$
            \State
            \State $\text{UpdateRewardParams}(a^*, r_t, y_t)$ 
            \State
            \State $A_i \gets A_i \setminus \{a_t^*\}$ \Comment{Remove chosen action.}
        \EndWhile
    \EndFor
\end{algorithmic}
\end{algorithm}

There are two limitations of the above learning strategy—First, the number of configurations $K$ can be potentially large--in experiments ID processing requires learning over nearly 500 choices! 
Second, the configuration-level does not account for the influence of individual operators on the overall performance of a configuration. We describe a novel hybrid strategy in which we determine the best configuration based on operator level estimates. 

\vspace{5pt}
\noindent {\bf Hybrid TS (\textbf{HTS}).} A drawback of CTS is that we may have a very large number of configurations, making it difficult to fully explore the configuration space. We would like to use the results at the operator level to inform our selection of which configuration to try next. For example, in ID processing, 
let’s say configuration $K_1 = [A_3, B_2, C_2]$, 
is the best operator choice configuration resulting in 10s of runtime but the assertion fails  for this configuration. Further, lets say we have also tried a few other configurations, such as:  $[A_3, B_2, C_1], [A_3, B_1, C_2],  [A_2, B_1, C_3], [A_4, B_1, C_2]$, 
and these configurations have also resulted in failure. Thus observing at the operator level, in all trials, operators $A_3$ failed thrice in total, $B_2$ failed twice, and $C_2$ failed thrice. However, we also know the entire configuration of $[A_3, B_2, C_2]$ also failed once. Even though we have only tried this configuration once, we have some idea of how it will perform based on the operators that make it up. Thus we will compute an empirical $\beta$ parameter (inspired by an empirical Bayes approach \cite{berger1985statistical}) for the configuration as: 
\begin{align*}
(3-1) + (2-1) + (3-1)  
\end{align*}
i.e. we sum the failures from individual operators, but we subtract the failure of the entire configuration to avoid double counting. Similarly, we also compute an empirical $\alpha$ parameter for the configuration based on successes. Finally, we set a hyperparameter $\gamma$ that influences how much to use the \operator{} level information when calculating our final $\alpha$ and $\beta$ parameters for the configuration. We ``allocate" $\gamma$ to the \job{} level parameters using a weighted average of successes and failures at the \operator{} level.

Formally, let $k$ be a \job{} in a workflow with $M$ operators. Let $\alpha’_k$ and $\beta’_k$ be the counts of successes and failure for the overall \textit{\job{}} $k$. Let $\alpha^{j}_k$ and $\beta^{j}_k$ be the counts of successes and failures for each \textit{\operator{}} $j$ in configuration $k$. We first calculate empirical Beta parameters for the configuration $k$, using our rule of summing the successes and failures for each operator $j$ that makes up configuration $k$, but subtracting off each time that the operator was used previously as part of configuration $k$ to avoid double counting: 
\begin{equation}
\label{eq:1}
    \alpha^{\textrm{em}}_k = \sum_{j=1}^{m} \left( \alpha^{j}_{k} - \alpha’_{k} \right), \quad   
    \beta^{\textrm{em}}_k = \sum_{j=1}^{m} \left( \beta^{j}_{k} - \beta’_{k} \right) \\
\end{equation}

We then calculate $\alpha_{k}$ and $\beta_{k}$, which are our hybrid parameters defining the reward distribution for \job{} $k$ under the hybrid strategy. For this, we sum the count of success and failures for \job{} $k$ ($\alpha’_k$ and $\beta’_k$) with our empirical prior but normalize the empirical prior to sum to one and control its influence using the hyperparameter $\gamma$. 

\begin{equation}
\label{eq:2}
    \alpha_{k} = \alpha’_{k} +\gamma \frac{\alpha^{\textrm{em}}_k}{\alpha^{\textrm{em}}_k + \beta^{\textrm{em}}_k}, \quad 
    \beta_{k} = \beta’_{k} + \gamma \frac{\beta^{\textrm{em}}_k}{\alpha^{\textrm{em}}_k + \beta^{\textrm{em}}_k} 
\end{equation}

The hybrid strategy updates parameters after processing each data item as is done previously in other strategies. It also chooses the next best configuration similar to the conditional rollback described earlier. From a pseudocode perspective, our hybrid strategy only changes the function \texttt{UpdateRewardParams} in Listing~\ref{alg:momc} which updates the Beta distribution parameters as described in Equations~\ref{eq:1} and \ref{eq:2}.  



\vspace{5pt}
\noindent {\bf Summary.} We summarize the different strategies for \toolname optimizers in \Cref{tab:ts_strategies}:
\begin{itemize}[leftmargin=*,noitemsep]
    \item \textbf{OTS} (Operator-level Thompson Sampling) maintains probability and runtime parameters at the operator level. This approach does not work for \momc{} workflows.
    \item \textbf{CTS} (Configuration-level Thompson Sampling) maintains parameters at both operator-level (to estimate runtime) and configuration-level (to estimate probability of passing). Actions are decided at the configuration level.
    \item \textbf{HTS} (Hybrid Thompson Sampling) maintains probability of passing estimates both at the operator and configuration-level and uses an empirical Bayes-inspired approach to allow operator estimates to inform configuration-level selections. 
\end{itemize}

OTS, HTS, and CTS are equivalent for \somc{} workflows.

\vspace{5pt}
\begin{table}[t]
    \centering
    \small
    \caption{Comparison of different Thompson Sampling based \toolname
    strategies. $C$ is the total count of operator choices across operators and
    $K$ is the count of configurations.}
    \vspace{-0.1in}
    \label{tab:ts_strategies}
    \begin{tabular}{
    |>{\centering\arraybackslash}m{1.0cm}|
    >{\centering\arraybackslash}m{.9cm}|
    >{\centering\arraybackslash}m{1.0cm}|
    >{\centering\arraybackslash}m{1.0cm}|
    >{\centering\arraybackslash}m{1.2cm}|
    >{\centering\arraybackslash}m{1.1cm}|
    }
        \hline
        {\textbf{Strategy}} & 
        {\textbf{Action Level}} &
        {\textbf{Operator Params}} & 
        {\textbf{Config Params}} & 
        {\textbf{Params Count}} & 
        {\textbf{Rollback-Aware}} \\\hline
        \multicolumn{6}{|c|}{Single operator with multiple choices} \\\hline
        OTS & \Operator{} & \shortstack{\\[0.2em]$\alpha_j, \beta_j$ \\ $\mu_j, \sigma_j^2$} &  & $O(C)$ &  \\
        \hline
        \multicolumn{6}{|c|}{Multiple operators with multiple choices} \\\hline
        CTS & \shortstack{\\[0.2em] Config} & $\mu_j, \sigma_j^2$ & $\alpha_k, \beta_k$ & $O(C+K)$ & \checkmark \\\hline
        HTS & \shortstack{\\[0.2em] Config} & \shortstack{\\[0.2em] $\alpha_j, \beta_j$ \\ $\mu_j, \sigma_j^2$} & \shortstack{\\[0.2em] $\alpha_k', \beta_k'$ \\ $\alpha_k^{em}, \beta_k^{em}$} & $O(C+K)$ & \checkmark \\\hline
    \end{tabular}
    \vspace{-0.1in}
\end{table}
\section{Evaluation}
\label{sec:experiments}

\vspace{5pt}
\noindent {\bf Implementation.}
\toolname implements operator choices, assertions, and
online optimizers in Python. \toolname is based on Scope~\cite{zhou2012scope} which provides a parallel dataflow language. It is written in $\sim$20k lines of Python code. Code and
experiment data presented in this paper is available~\cite{popper}. 

A choice is programmatically expressed within an operator as follows: 
\vspace{5pt}
\begin{lstlisting}[language=Python]
class OCR(Processor):
  def __call__(self, input_row: Row, ...):
    img = input_row["IMG"]
    yield Row({"OCR": [lambda img=img: easy_ocr(img), 
                lambda img=img: tesseract(img)]}, ...)
\end{lstlisting}
\vspace{5pt}

In this implementation, the OCR operator encapsulates two possible choices—EasyOCR and Tesseract—which are represented as lambda functions. 
This means that either EasyOCR or Tesseract can be instantiated at any given
time. Assertions are specified using functions that return \texttt{False} in
case of an error, \texttt{True} otherwise. For example, the following assertion
checks that the name should not be empty and should start with an upper case
character:

\vspace{5pt}
\begin{lstlisting}[language=Python]
def check(self, input_row: Row):
  name = input_row["name"]
  if len(name) == 0 or not name[0].isupper():
    return False, "invalid name"
  return True, None
\end{lstlisting}
\vspace{5pt}

While building a workflow, users can specify the optimizer.
A complete workflow using operators 
, assertions, and optimizer
is shown below. This workflow is instantiated with the HTS optimizer.

\vspace{5pt}
\begin{lstlisting}[language=Python]
jb = JobBuilder(HTS(..))
n = jb.extract(ImgReader(..))
      .process(ChannelPicker(..))
      .process(OCR(..))
# ...
job = jb.union(inps=[alb, esp, ...])
        .assrt(predicate=check)
        .output(CsvWriter(..))
        .build(..)
\end{lstlisting}
\vspace{5pt}

All observed metadata such as runtimes corresponding to operator choices and success and failures of the assertion are maintained in Redis. Redis also maintains the ancestor row lineage information. The rollback uses row-level lineage and employs fast incremental computation  described in more detail in~\cite{deem24reactive}.



\newsavebox{\cards}
\savebox{\cards} {
\begin{tikzpicture}[node distance=20pt]
\node[] (ListDir) {ID card image~\cite{midv2020}};
\node[draw, below of=ListDir] (Crop) {Crop};
\node[draw, below of=Crop, align=center, yshift=-10pt] (Contrast) {\textbf{Contrast}\\original, increased};
\node[draw, below of=Contrast, align=center, yshift=-10pt] (Sharpness) {\textbf{Sharpness}\\original, increased};
\node[draw, below of=Sharpness, align=center, yshift=-10pt] (ColorBalance) {\textbf{Color Balance}\\original, decreased};
\node[draw, below of=ColorBalance, align=center, yshift=-20pt] (ChannelPicker) {\textbf{Channel Picker}\\original, red, green, blue, \\red+green, red+blue, blue+green};
\node[draw, below of=ChannelPicker, align=center, yshift=-20pt] (OCR) {\textbf{OCR}\\Tesseract, EasyOCR};
\node[draw, below of=OCR, align=center, yshift=-10pt] (Classifier) {\textbf{Classifier}\\class-1, class-2};

\node[draw, below of=Classifier, xshift=-26pt, yshift=-10pt] (fESP) {$\sigma_{class=ESP}$};
\node[draw, below of=fESP] (ESP) {ESP Extract};
\node[draw, below of=Classifier, xshift=26pt, yshift=-10pt] (fALB) {$\sigma_{class=ALB}$};
\node[draw, below of=fALB] (ALB) {ALB Extract};
\node[below of=Classifier, yshift=-10pt] (extra) {...};

\node[draw, below of=Classifier, yshift=-50pt] (Union) {Union};
\node[draw, below of=Union] (Assert) {date of birth < date of issue $\land \cdots$};
\node[below of=Assert] (CsvWriter) {first name, last name, dob, ..};
\node[below of=CsvWriter, align=center] (Caption) {\textbf{(d) ICP: ID card}\\\textbf{processing (224 configs)}};

\draw[->] (ListDir) -- (Crop);
\draw[->] (Crop) -- (Contrast);
\draw[->] (Contrast) -- (Sharpness);
\draw[->] (Sharpness) -- (ColorBalance);
\draw[->] (ColorBalance) -- (ChannelPicker);
\draw[->] (ChannelPicker) -- (OCR);
\draw[->] (OCR) -- (Classifier);

\draw[->] (Classifier) -- (fESP);
\draw[->] (fESP) -- (ESP);
\draw[->] (Classifier) -- (fALB);
\draw[->] (fALB) -- (ALB);

\draw[->] (ESP) -- (Union);
\draw[->] (ALB) -- (Union);
\draw[->] (Union) -- (Assert);
\draw[->] (Assert) -- (CsvWriter);

\draw[->, dashed, bend left=65] (Assert) to (Contrast);
\draw[->, dashed, bend left=70] (Assert) to (Sharpness);
\draw[->, dashed, bend left=70] (Assert) to (ColorBalance);
\draw[->, dashed, bend left=70] (Assert) to (ChannelPicker);
\draw[->, dashed, bend left=90] (Assert) to (OCR);
\draw[->, dashed, bend left=90] (Assert) to (Classifier);
\end{tikzpicture}
}

\newsavebox{\rag}
\savebox{\rag} {
\begin{tikzpicture}[node distance=20pt]
\node[] (LineReader) {Question, Context (with distractions) ~\cite{yang2018hotpotqa}};

\node[draw, below of=LineReader, align=center, yshift=-40pt] (ModelPicker) {\textbf{Embedding Model}\\all-MiniLM-L6-v2,\\all-mpnet-base-v2,\\all-MiniLM-L12-v2,\\bge-m3,\\paraphrase-multilingual-MiniLM-L12-v2,\\paraphrase-multilingual-mpnet-base-v2,\\LaBSE};

\node[draw, below of=ModelPicker, align=center, yshift=-55pt] (VectorStore) {\textbf{VectorStore}\\(350, 8), (400, 7), (560, 5),\\(700, 4), (1400, 2)};

\node[draw, below of=VectorStore, align=center, yshift=-20pt] (Summarizer) {\textbf{Retriever}\\similarity, mmr};

\node[draw, below of=Summarizer, align=center, yshift=-50pt] (QA) {\textbf{QA Model}\\tinyroberta,\\distilbert-base-cased,\\roberta-base,\\mobilebert-uncased,\\google-bert,\\minilm-uncased,\\xlm-roberta-large};

\node[draw, below of=QA, yshift=-50pt] (Assert) {confidence > 0.65};

\node[below of=Assert] (CsvWriter) {Answer};
\node[below of=CsvWriter, align=center] (Caption) {\textbf{(c) LCQA: Large-context}\\\textbf{quest.\ answer.\ (490 configs)}};

\draw[->] (LineReader) -- (ModelPicker);
\draw[->] (ModelPicker) -- (VectorStore);
\draw[->] (VectorStore) -- (Summarizer);
\draw[->] (Summarizer) -- (QA);
\draw[->] (QA) -- (Assert);
\draw[->] (Assert) -- (CsvWriter);

\draw[->, dashed, bend left=50] (Assert) to (ModelPicker);
\draw[->, dashed, bend left=65] (Assert) to (VectorStore);
\draw[->, dashed, bend left=65] (Assert) to (Summarizer);
\draw[->, dashed, bend left=70] (Assert) to (QA);
\end{tikzpicture}
}




\newsavebox{\qa}
\savebox{\qa} {
\begin{tikzpicture}[node distance=20pt]
\node[] (Image) {Image~\cite{deng2009imagenet}};
\node[draw, below of=Image, align=center, yshift=-18pt] (Classifier) {\textbf{Classify}\\DeiT tiny~\cite{pmlr-v139-touvron21a},\\ViT~\cite{wu2020visual}};
\node[draw, below of=Classifier, yshift=-10pt] (ClassAssert) {confidence > 0.65};
\node[below of=ClassAssert] (ClassCsvWriter) {Class};
\node[below of=ClassCsvWriter, align=center] (ClassCaption) {\textbf{(a) IC: Image}\\\textbf{classification (2 configs)}};

\draw[->] (Image) -- (Classifier);
\draw[->] (Classifier) -- (ClassAssert);
\draw[->] (ClassAssert) -- (ClassCsvWriter);

\draw[->, dashed, bend left=80] (ClassAssert) to (Classifier);

\node[below of=ClassCaption, yshift=-90pt] (LineReader) {Question, Context~\cite{squad}};
\node[draw, below of=LineReader, align=center, yshift=-18pt] (QA) {\textbf{QA}\\TinyBert~\cite{TinyBERT},\\RoBERTa~\cite{roberta}};
\node[draw, below of=QA, yshift=-10pt] (Assert) {confidence > 0.65};
\node[below of=Assert] (CsvWriter) {Answer};
\node[below of=CsvWriter, align=center] (Caption) {\textbf{(b) SCQA: Small-context}\\\textbf{quest.\ answer.\ (2 configs)}};

\draw[->] (LineReader) -- (QA);
\draw[->] (QA) -- (Assert);
\draw[->] (Assert) -- (CsvWriter);

\draw[->, dashed, bend left=80] (Assert) to (QA);




\end{tikzpicture}
}

\begin{figure}[t]
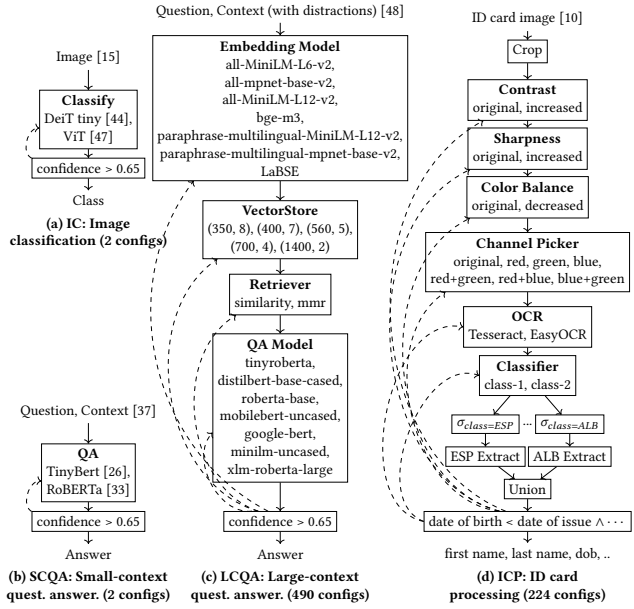

  \setlength\belowcaptionskip{-12pt}
  \resizebox{\linewidth}{!}{
    \centering
    \subfloat{\usebox{\qa}}\hspace{-45pt}
    \subfloat{\usebox{\rag}}\hspace{-45pt}
    \subfloat{\usebox{\cards}}
  }
  \caption{Real ML workflows for evaluating \toolname.}
  \label{fig:apps}
\end{figure}

\vspace{5pt}
\noindent {\bf Experiment setup.} We run all the experiments on a server with Intel Xeon Gold 6330 and an
Nvidia A40 GPU. 
We evaluate whether the optimizers proposed in this paper (OTS, CTS,
HTS in Table~\ref{tab:ts_strategies}) can effectively lower runtimes.
We perform an ablation study by selectively removing learning capabilities from
the proposed optimizers. In particular, we ablate OTS to OTS-R, which does not model operator runtimes ($\mu_a, \sigma_a$ in Algorithm~\ref{alg:somc}); it only learns the probability that each operator choice passes the assertion and
assumes equal runtimes for all choices. We ablate CTS to CTS-RB, which is not rollback-aware and selects the next configuration without considering the transition cost from the previously attempted one, \ie \texttt{SelectAction} in Algorithm~\ref{alg:momc}
does not consider $a^*_{t-1}$.

\noindent {\bf ML Workflows.} We evaluate \toolname using four real-world ML workflows and two synthetic workflows as shown in Figure~\ref{fig:apps} and Figure~\ref{fig:micro}. 

\noindent {\bf Real workflows.} The four real-world workflows are chosen from
popular online sources described below. Each workflow includes the three core constructs—operator choices,
assertions, and rollbacks—as illustrated in Figure~\ref{fig:apps}. 
The original versions of these workflows did not include
published operator choices. Instead, each configuration involving a specific
model operator was published independently. 
The real input datasets used by the workflows are cited at the
beginning of each, while the type of output produced is described at the end.

\vspace{5pt}
\noindent {\bf \uline \ic.} This ML workflow is based on the HuggingFace image
classification pipeline. The workflow processes images from the
ImageNet~\cite{deng2009imagenet} dataset. There is a single classify operator
that defines two model choices: a lightweight DeiT
tiny~\cite{pmlr-v139-touvron21a} model and a heavier ViT~\cite{wu2020visual} model. The
assertion checks if the confidence of the model prediction is above a specified threshold. 

\vspace{5pt}
\noindent {\bf \uline \scqa.} This ML workflow is structurally similar to the previous
one and is based on the HuggingFace question answering pipeline. It answers
questions from the SQuaD~\cite{squad} dataset containing a single QA operator
that defines two model choices: a lightweight TinyBERT~\cite{TinyBERT} model and a
heavier RoBERTa~\cite{roberta} model.  The assertion is similar to the one in \ic. Our
experiments will reveal that even though these workflows are structurally similar, \toolname
strategies can learn different optimal sequences of configurations which depends
on runtimes and probabilities of passing assertions for different choices.


\vspace{5pt}
\noindent {\bf \uline \lcqa.} This workflow is for multi-hop question answering and is based on the popular benchmark HotpotQA, in the "distractor"  mode~\cite{yang2018hotpotqa}. Briefly, each question is given paragraphs from 10 different Wikipedia entries as context. Exactly two of the paragraphs contain the necessary \emph{supporting facts} that together allow one to determine the answer.\footnote{Having to analyze facts across two separate paragraphs is what makes this task "multi-hop."} The remaining paragraphs are deliberate distractions.  The answers are usually one to three words long, and each predicted answer is evaluated against the ground truth using the $F_1$ metric.\footnote{\emph{Precision} is defined as the ratio of the number of tokens common to both the prediction and the ground truth to the number of tokens in the prediction.  \emph{Recall} and $F_1$ are defined similarly.  See, e.g., \cite{wikipedia_precision_recall_2025}.}  Our workflow is similar to a standard RAG-based workflow using the LangChain orchestration framework (see, e.g., \cite{langchain_rag_tutorial2025}).  Its essential components are as follows: All the provided paragraphs are first concatenated and then split into chunks.  An embedding vector is generated for each chunk using one of 7 embedding models (the first operator; see Figure \ref{fig:apps}(c) for model choices). The size of each chunk is chosen from 5 different options (second operator; called type of VectorStore).  The embeddings of the question and the chunks are then used to retrieve the most relevant chunks, again using one of two retrieval mechanisms (third operator).  The number of chunks retrieved can also vary, but for implementation reasons, this is a fixed number for each VectorStore option.  So, e.g., the VectorStore option $(350, 8)$ is the option in which the size of each chunk is $350$ characters and $8$ of the most relevant chunks are returned by the retriever.  The relevant chunks are fed into one of $7$ question-answering (QA) models (fourth and last operator).\footnote{Our models are general purpose QA models commonly used in HuggingFace/LangChain and not specially fine-tuned to identify supporting facts in the HotPotQA dataset.}  Finally, our assertion is that the confidence score computed by the QA model should meet a minimum threshold, which was set to $0.65$ in our experiments.
This workflow has a total of $490$ configurations. 

\vspace{5pt}
\noindent {\bf \uline  \icp.} This ML workflow is inspired from articles describing KYC
pipelines~\cite{icp3} deployed in the industry and open KYC Github
repositories~\cite{icp_github, icp_github2}. This workflow extracts fields from
ID card scans from the MIDV-2020 dataset~\cite{midv2020}. It crops the images to
make them upright and then does various image cleaning operations like changing
contrast, sharpness, color balance, and picks color channels to make the images
more amenable to OCR. Then, it uses OCR engines and country-specific field
extractor models to extract fields like first name, last name, and date of
birth. Extracted fields are evaluated against ground truth using the $F_1$
metric.

Image cleaning operations are made optional by keeping the original image as one of the operator choices. The OCR
operator specifies two different OCR engines as choices. The classifier operator
specifies top-2 predictions from a ResNet~\cite{resnet2015} classifier as
choices. The assertions verify several data integrity constraints, such as ensuring that the extracted name is not null and that the extracted date of birth precedes the extracted date of issue.
This workflow has a total of $224$ configurations.




\newsavebox{\microsimpledelay}
\savebox{\microsimpledelay} {
\begin{tikzpicture}[node distance=50pt]
\node[] (Input) {};
\node[draw, right of=Input, align=center] (Choice) {\textbf{Choice}\\c = 1 (0.15s),\\2 (0.2s), 3 (0.2s),\\ 4 (0.5s), 5 (1s)};
\node[draw, right of=Choice, align=center, xshift=20pt] (Assert) {\textbf{Assertion}\\p=0.16*c};
\node[below=of Choice, align=center, xshift=30pt, yshift=45pt] (Caption) {\textbf{(a) \snd: \somc~(5 configs)}};

\draw[->] (Input) -- (Choice);
\draw[->] (Choice) -- (Assert);

\draw[->, dashed, bend left=40] (Assert) to (Choice);
\end{tikzpicture}
}




\newsavebox{\micromultidelay}
\savebox{\micromultidelay} {
\begin{tikzpicture}[node distance=50pt]
\node[] (Input) {};
\node[draw, right of=Input, align=center] (Choice) {\textbf{Choice 1}\\c1 = 0(0.2s),\\ 1(0.1s), 2(0.2s),\\ 3(0.2s), 4(1s)};
\node[draw, right of=Choice, align=center, xshift=20pt] (Choice2) {\textbf{Choice 2}\\c2 = 0(0.2s),\\1(0.1s), 2(0.2s),\\3(0.2s), 4(1s)};
\node[draw, right of=Choice2, align=center, xshift=20pt] (Assert) {\textbf{Assertion}\\p=0.8*(c1$\oplus$c2)/7};
\node[below of=Choice2, align=center, yshift=5pt] (Output) {\textbf{(b) \mnd: \momc~(25 configs)}};
\draw[->] (Input) -- (Choice);
\draw[->] (Choice) -- (Choice2);
\draw[->] (Choice2) -- (Assert);

\draw[->, dashed, bend left=40] (Assert) to (Choice);
\draw[->, dashed, bend left=40] (Assert) to (Choice2);
\end{tikzpicture}
}

\begin{figure}[t]
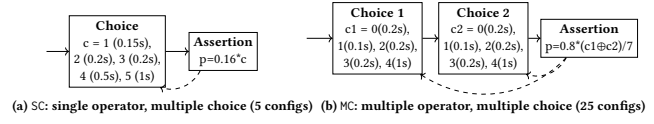

  \resizebox{\linewidth}{!}{
    \centering
    \hspace{-10pt}
    \subfloat{\usebox{\microsimpledelay}}\hspace{-20pt}
    \subfloat{\usebox{\micromultidelay}}
  }
  \vspace{-0.3in}
  \caption{Synthetic workflows for evaluating \toolname strategies.}
  \vspace{-0.2in}
  \label{fig:micro}
\end{figure}

\vspace{5pt}
\noindent {\bf Synthetic workflows.} Since we do not know the optimal sequence
of configurations for real-world ML workflows, we also evaluate \toolname
strategies with synthetic workflows, as illustrated in Figure~\ref{fig:micro}, so that we can compare \toolname performance to an optimal strategy.
We calculate each workflow's optimal sequence of configurations,
based on Equation~\ref{eq:sequence_runtime}, to evaluate whether and how
quickly \toolname strategies learn to provide optimal runtimes. Each
synthetic workflow contains one or two choice operators, a single assertion, and
the ability to rollback to any choice operator.

The pass probability of the assertion depends on the selected choice/(s). In \snd,
the probability of passing the assertion is kept as $0.16c$, where $c$ is the
choice index. Therefore, even the best choice has a probability of $0.8$ \ie it
does not always pass the assertion. This mimics the behavior of models in real world
ML workflows.

Each choice is associated with a
runtime cost. For instance, choice $c = 1$ in \snd incurs a runtime of 0.15
seconds. 
We chose these values to mimic real world situations where heavier models are
often more accurate but also take more time. For example, choice 5 has a
higher probability (0.80) of passing the assertion than choice 4 (0.64), but
choice 5 takes double the time (1s) to run than the choice 4 (0.5s).

For \mnd, the assertion is based on the combination of
choice index values. We again keep the maximum probability as $0.8$ so that none
of the configurations always pass the assertion.
To stress test the learning task of \toolname strategies, we keep the choices in
the two operators $c1$ and $c2$ dependent on each other by setting the
probability  $0.8 \times (c1\;\oplus\;c2)/7$, where $\oplus$ denotes the bitwise XOR
operation. For example, $c1=4$ has a $0.8$ probability of passing assertions
when $c2=3$ but has a $0$ probability with $c2=4$.

As shown in the figure, each choice operator
provides 5 possible choices. While the number of configurations can be varied, we
kept a maximum of 25 configurations so that the optimal sequence of configurations
could be computed by a brute-force approach with sequence pruning. We omit the
details of this brute-force approach due to space limitations. Prior knowledge of
optimal ordering allows us to evaluate whether and how quickly \toolname
strategies converge to the same runtime as the optimal strategy. We also ran our experiments with
other runtimes and pass probabilities in the synthetic workflows.  Since the overall results were not much different, we skip them here due to space constraints.


\vspace{5pt}
\noindent {\bf Evaluation Goals and Baselines.} We evaluate \toolname on three fronts: 
\begin{enumerate}[leftmargin=*,noitemsep]
\item Do \toolname's learning strategies quickly learn optimal runtimes for
diverse ML workflows? This evaluation helps to assess the learning rate of the
single operator and multi operator learning strategies. We evaluate this
question using both synthetic and real workflows. In synthetic workflows, the
optimal sequence serves as the baseline for comparison. In real workflows, since
 the optimal sequence is not available, we assess how quickly does a strategy converge  to stable regret in
the presence of a large number of configurations \ie determine the scalability of
strategies. We also present case studies to show how optimizers learn efficiently even
in workflows with hundreds of configurations.  (Section~\ref{sec:perf})


\item Do \toolname's programming abstractions--operator choices and
assertions--improve accuracy compared to the single best configuration? We
evaluate this on real ML workflows by separately running each workflow with
individual configurations and comparing each configuration's accuracy with
the accuracy provided by \toolname. (Section~\ref{sec:accuracy})
\tm{I agree with Todd. Do not understand this para.} 

\item Finally, how does \toolname's performance compare with other
similar dataflow systems? For this goal, we use real workflows and compare their
performance on \toolname and state-of-the-art dataflow systems Spark~\cite{zaharia2012spark} and
Flink~\cite{carbone2017flink}. (Section~\ref{sec:runtime})
\end{enumerate}

\subsection{Performance of \toolname optimizers}
\label{sec:perf}
\tm{This section is performance but I think given the question above we should call it learning Rate of \toolname optimizers}
In this section, we evaluate whether the optimizers proposed in this paper (OTS,
CTS, HTS in Table~\ref{tab:ts_strategies}) can lower runtimes for
both synthetic and real-world ML workflows. We perform ablation studies by
also evaluating OTS-R (OTS without runtime modeling) and CTS-RB (CTS without
rollback awareness). 


\subsubsection{Performance of optimizers on synthetic workflows}
\tm{this use of time series is a bit confusing...}
For each synthetic workflow, we first run it multiple times with its
optimal sequence of configurations to get a series of optimal runtimes.
Due to probabilistic assertions, choices may fail and cause rollbacks.
Hence, the runtimes vary across different runs. Next, with a given optimizer,
we run the workflow multiple times, with the optimizer updating itself after each
run, to get a series of optimizer runtimes. We then compare the series
of optimizer runtimes against that of optimal runtimes. Since runtimes vary
across each run, we 
show runtime trends by 
plotting a sliding-window
runtime average. In addition, we plot the cumulative regret curves
for each optimizer. Cumulative regret is measured as the cumulative difference
between the achieved runtime and the optimal runtime. Plotted as a series,
cumulative regret curves show how fast an optimizer approaches the optimal
runtime; a good learning strategy will show a ``log-like shape" in its cumulative regret
curve with a rapidly decreasing slope. 
Note that because of probabilistic assertions, learning strategies can sometimes get
lucky and finish faster than the optimal strategy, leading to negative regret.

\begin{figure}[t]
    \subfloat[\snd]{\label{fig:simple_with_delay}\includegraphics[width=0.48\linewidth]{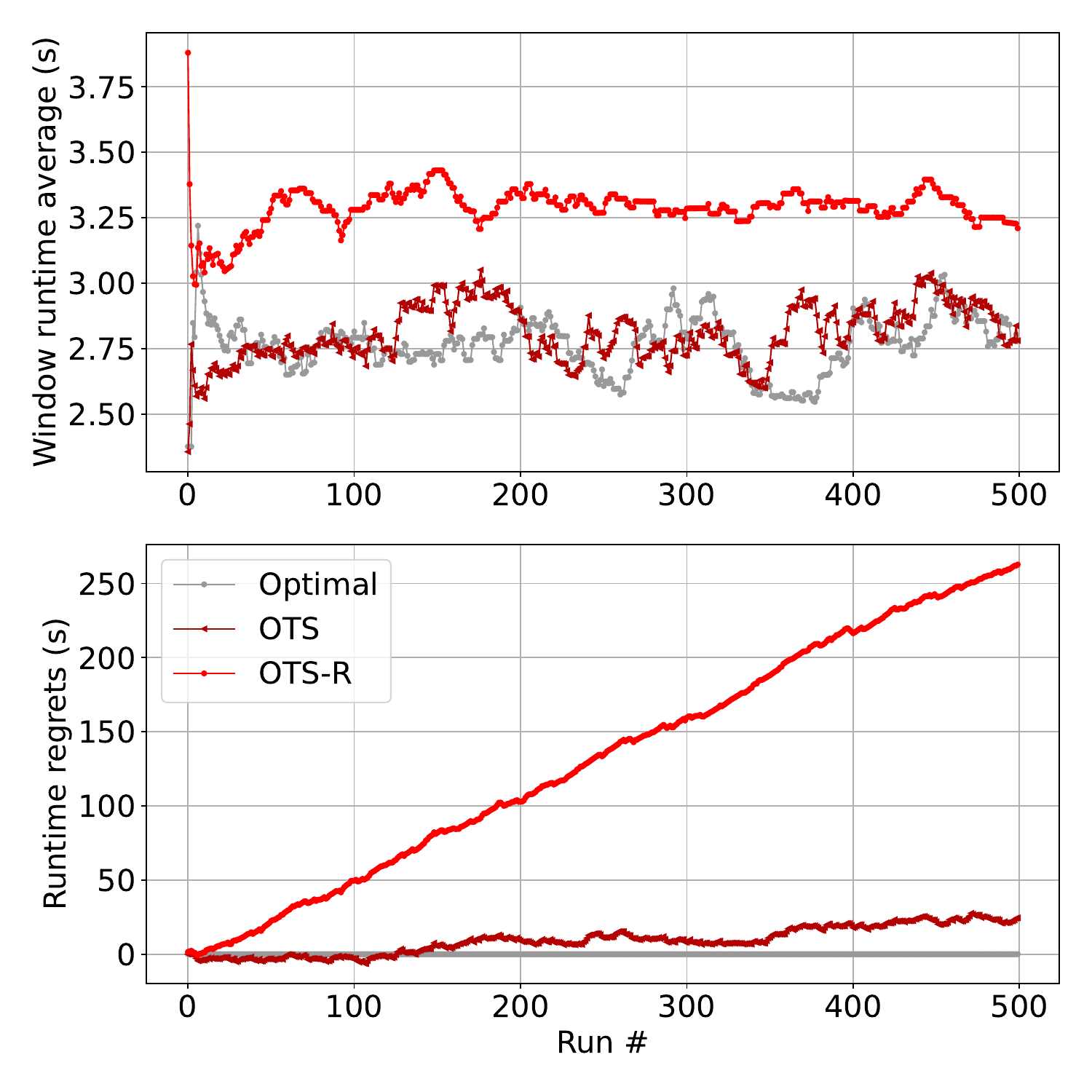}}
    \subfloat[\mnd]{\label{fig:multiple_with_delay}\includegraphics[width=0.48\linewidth]{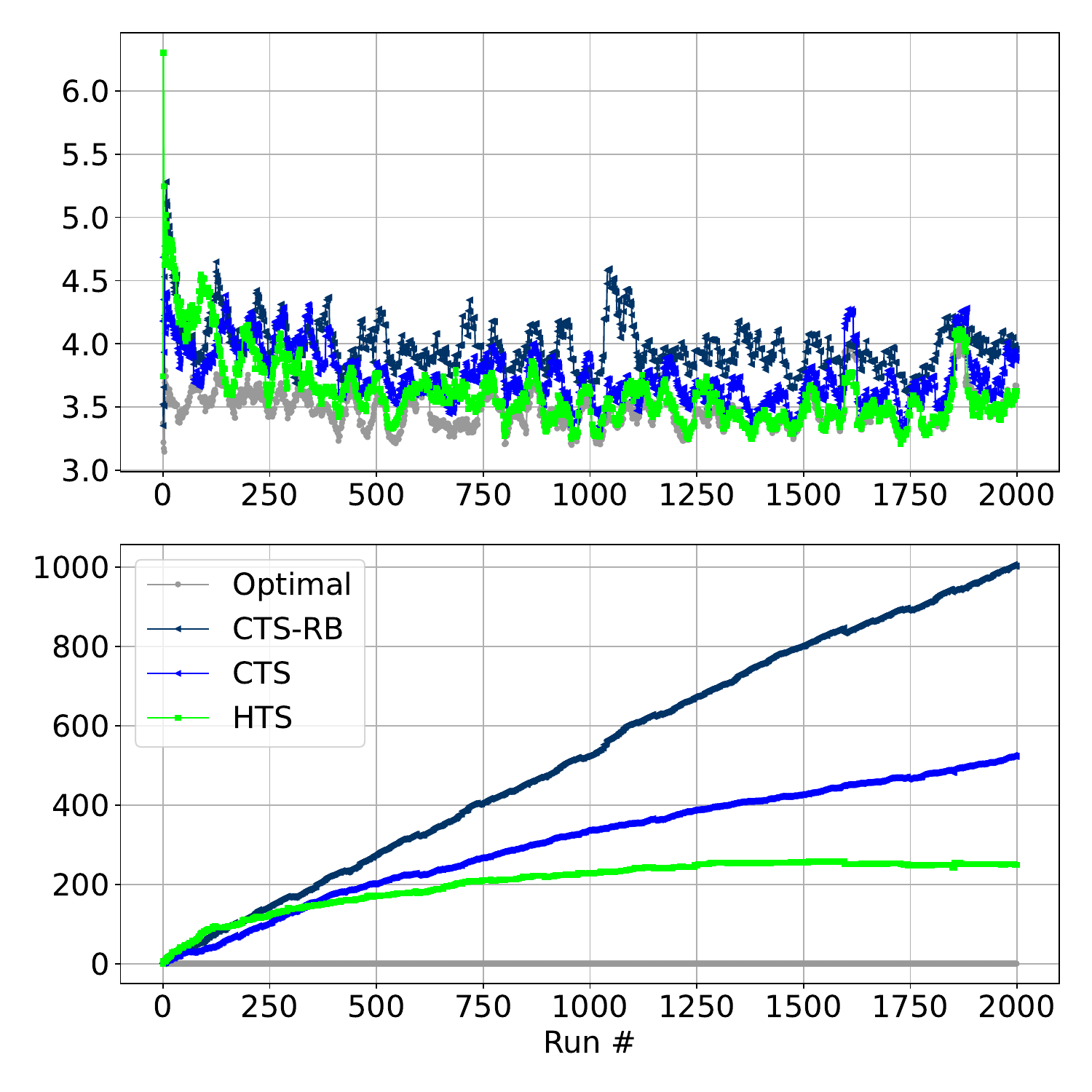}}
  \vspace{-0.1in}
  \caption{Modeling runtime and being rollback aware is important for strategies to learn.}
  \vspace{-0.1in}
\end{figure}


\paragraph{\snd} 
We evaluate OTS and OTS-R for \snd.
We see from Figure~\ref{fig:simple_with_delay} that the
runtime-agnostic OTS-R is unable to reach the optimal runtime. This is because
OTS-R simply maximizes probability of passing assertions. It learns to run the
sequence $c=[5,4,3,2,1]$, but the optimal sequence for \snd is $c=[3, 2, 4, 1, 5]$
due to different runtimes of choices. The runtime-aware multi-armed bandit OTS
is able to learn the optimal order of choices maintaining a much lower regret
than OTS-R. 



\paragraph{\mnd}
We next evaluate CTS, HTS, and CTS-RB for \mnd.
From Figure~\ref{fig:multiple_with_delay}, we observe that the CTS-RB, 
that does not take rollback order into account, cannot learn the optimal
order. This is because this learning strategy just picks configurations with the
lowest expected runtime without considering the computational structure of the
workflow. For example, in the last run, CTS-RB tries $(c_1=3, c_2=4)$
followed by $(c_1=4, c_2=2)$. 
In the 
same run, both CTS and HTS try $(c_1=4, c_2=1)$ followed by $(c_1=4, c_2=0)$.
Switching from $(c_1=4, c_2=1)$ to $(c_1=4, c_2=0)$ 
avoids re-executing
$c_1=4$. This shows that rollback awareness is essential for learning in
workflows with multiple choice operators.

Further, we observe in Figure~\ref{fig:multiple_with_delay} that HTS learns much faster than CTS, resulting in lower cumulative
regret. This implies  in HTS augmenting the probability estimations for configurations
with the estimations for individual choices makes learning more effective.



\noindent\textit{Our key observations are as follows: (i) accurately modeling both runtime and the probability of assertion success is crucial for determining the optimal execution order; (ii) while the OTS strategy is effective for workflows involving a single choice operator, workflows with multiple choice operators require strategies that are aware of potential rollbacks; and (iii) the hybrid strategy HTS—combining configuration-level learning with operator-level insights—achieves faster convergence compared to the configuration-only CTS approach.}

\subsubsection{Performance of optimizers on real-world ML workflows}
\label{sec:applications_experiments}

In this section, we shift our focus to real-world workflows and analyze how these strategies impact runtime performance. Specifically, we determine the effect of OTS on 
on runtimes of single operator workflows, \ic and \scqa, 
and the impact of HTS on runtimes of multiple operator workflows, \lcqa and \icp.

Each workflow is executed with 1 input per run. After each run, we
unload all the ML models from the GPU memory and stop all workers. Since, the optimal sequence of configurations
for these real workflows is unknown, we evaluate how quickly learning strategies converge by reporting
their cumulative runtime average. We also analyze the final configuration sequence 
learned by optimizers after convergence.

\noindent {\it OTS Convergence Behavior on Real Workflows.} OTS learns effectively within the first 50 runs and consistently maintains low runtime for both \ic and \scqa (figures omitted due to space constraints).  For \ic, OTS learns to favor the lightweight DeiT-Tiny model over the heavier ViT model. DeiT-Tiny loads and runs inference in 2.0 seconds, compared to 2.4 seconds for ViT, and satisfies the assertion in 82\% of cases—making it a strong first choice.
However, for \scqa, OTS learns
to prioritize the heavier RoBERTa model over the lighter TinyBERT. The optimizer
observes that the runtime difference is minimal—1.4s for TinyBERT versus 1.5s
for RoBERTa. Moreover, the lightweight model often fails to satisfy the
assertion, making the heavier model the better choice.

\begin{figure}[t]
  \subfloat[LCQA]{\label{fig:learn_lcqa}\includegraphics[width=0.5\linewidth]{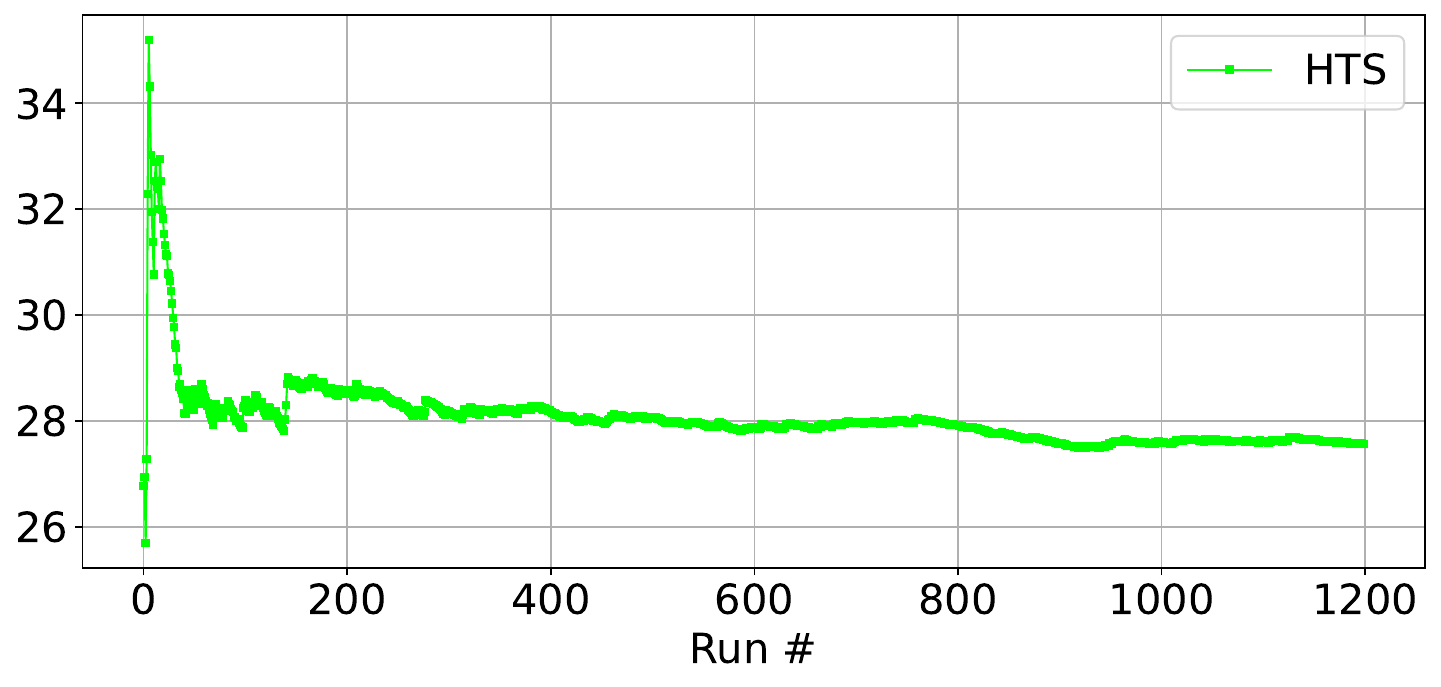}}
  \subfloat[Cards]{\label{fig:learn_cards}\includegraphics[width=0.5\linewidth]{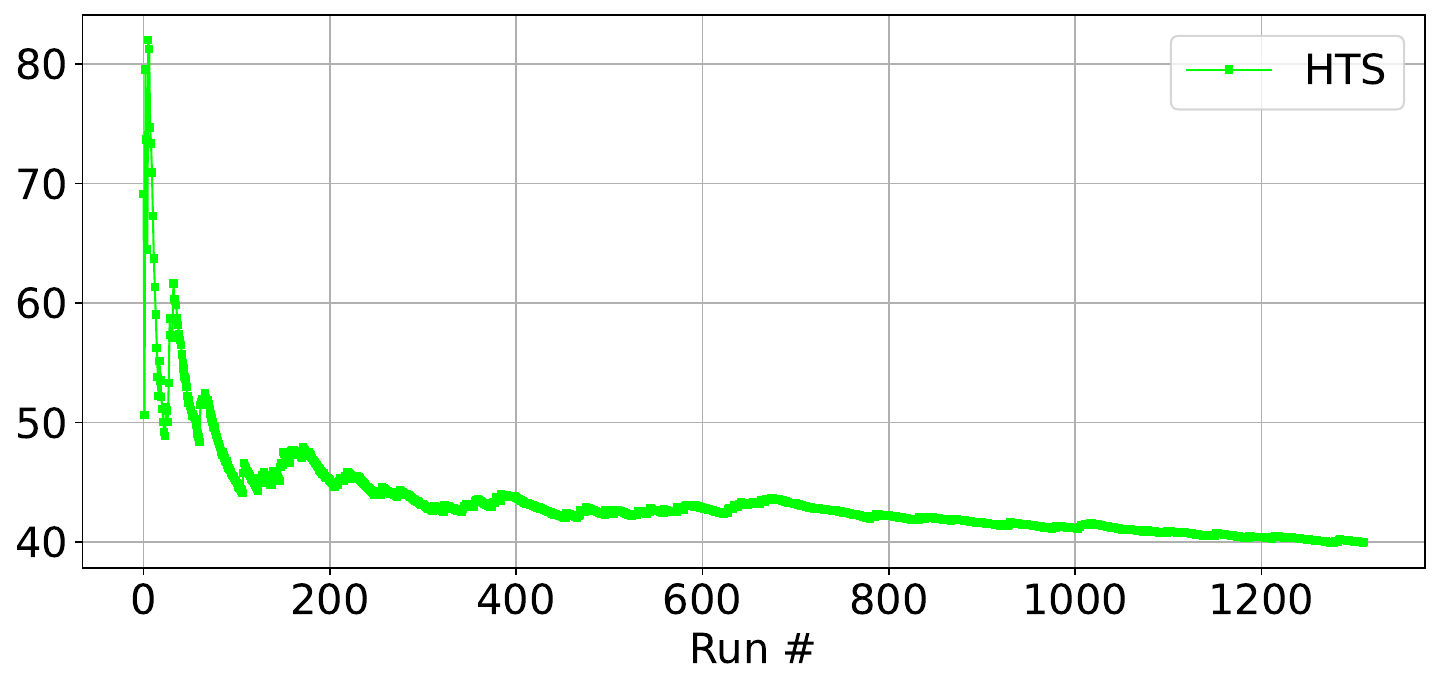}}
  \vspace{-0.1in}
  \caption{\toolname optimizers can effectively improve runtimes for real-world ML workflows.}
  \vspace{-0.2in}
\end{figure}

\noindent {\it HTS Convergence Behavior on Real Workflows.} Figure~\ref{fig:learn_lcqa} shows that the HTS runtime for the \texttt{LCQA} workflow converges after about
1000 runs. In its final run, the optimizer selects
the configuration <all-MiniLM-L6-v2, (700, 4), mmr, google-bert>\footnote{Recall a
configuration for \lcqa specifies operator choices for $\langle$Embedding Model,
VectorStore chunk size and number of chunks retrieved, Retriever, QA
Model$\rangle$; also see Figure \ref{fig:apps} (c).}. Interestingly, this choice is not the one with the highest scores---its $\alpha$ = 36.4 \footnote{Recall that HTS
sums up a fraction of $\alpha$ and $\beta$ values from the operator level. They
may therefore not be an integer.} and the empirical mean of 0.46, rank only third among the 490 configurations. However, because HTS is rollback
aware, it explores other potential configurations probabilistically.

Thus, after this initial configuration, the optimizer systematically explores the configurations as follows: 
(i) it tests 6 different QA models; (ii) it changes the retriever from mmr to similarity in the next 7 configurations; (iii) it runs with the chunk size 400 for the next 14 configurations, and, finally (iv) it changes the embedding model to LaBSE.


\noindent {\it HTS Converges Similarly but Learning Behavior Differs in the \icp Workflow.} Figure~\ref{fig:learn_cards} shows HTS also 
improves runtimes for the
\icp workflow, though the learned configuration sequence 
behaves quite differently from \lcqa. In \lcqa, multiple  operators contribute
non-trivially to the total runtime. But in \icp, only the OCR operator has a substantial
runtime; all the other choices take less than 80ms. Further,
it learns that Tesseract passes 9 times less often than EasyOCR. Consequently, 
instead of following a bottom-up strategy as in \lcqa, HTS adapts to the runtime profile of \icp as follows: It delays trying Tesseract until the 17th  configuration in the final sequence; The first configuration disables most image-cleaning steps; When that fails, it tries the next prediction from the classifier since it does not
need another model inference; Next, it tries modifying sharpness and later switches to a red+green color channel.



\textit{Overall, we find that \toolname optimizers learn efficiently even in workflows with hundreds of configurations. By leveraging assertion success rates, operator runtimes, and workflow structure, the optimizers adapt their strategies—favoring lightweight models when feasible and selecting more reliable ones when necessary. }





\vspace{-5pt}
\subsection{Accuracy of \toolname} 
\label{sec:accuracy}
\begin{figure}[t]
	\centering
    \includegraphics[width=0.8\linewidth]{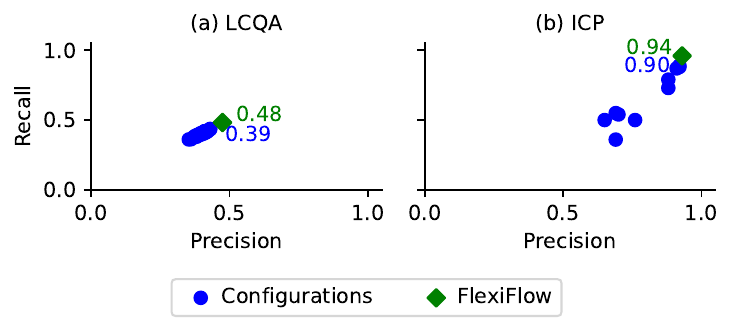}
    \vspace{-0.1in}
    \caption{Effectiveness of using operator choices in ML workflows.
    \toolname achieves a higher $F_1$ score than the best configuration.}
    \vspace{-0.2in}
    \label{fig:effective}
\end{figure}

We now examine how and why \toolname delivers more accurate results by
considering the question answering workflow \lcqa and the document processing workflow
\icp. We compare the accuracy of \toolname with that of individual workflow
configurations run separately on identical input. Instead of individually running
the large number of configurations, we run a reduced version of the workflows. For
\lcqa, we use two of the seven embedding models: all-MiniLM-L6-v2 and LaBSE, and
two of the seven QA models: xlm-roberta-large and tinyroberta, totaling 40
configurations. For \icp, we remove contrast, sharpness, and color balance image
cleaning operators, and remove red+blue and red+green options from
ChannelPicker, totaling 20 configurations (refer Figure~\ref{fig:apps}(d)). \tn{I'm not sure what this last sentence means:} \toolname version also runs these
reduced workflows.

\toolname achieves an $F_1$ score of $0.48$ (see 
Figure~\ref{fig:effective} (a)), a relative improvement of 23\% over the best single underlying configuration---$\langle$all-MiniLM-L6-v2, (350, 8), similarity, xlm-roberta-large$\rangle$, which only achieves a score of $0.39$.\footnote{We chose general purpose QA Models rather than ones specially fine-tuned for HotpotQA because our primary objective here is to evaluate the performance of \toolname w.r.t.\ a given a set of configurations.  As such, our models score lower than the state-of-the-art for this dataset.} As the figure shows, \toolname is better in both precision and recall.

The reason for the significant improvement in performance is that
different underlying configurations are \emph{uncorrelated}---they differ in the
examples they get wrong. For instance, Figure~\ref{fig:venn} shows that the
configuration with the best $F_1$ metric gets 174 questions correct (using the \emph{exact match} metric).
Among the remaining questions, a different configuration $\langle$LaBSE,
(560,5), similarity, tinyroberta$\rangle$ gets 169 questions correct.
\footnote{Observe that the former configuration had the best $F_1$ score, while the latter configuration had a higher exact match score.}
\def\firstcircle{(0,0) circle (1.74cm)} 
\def\thirdcircle{(0:-0.6cm) circle (2.29cm)} 
\newsavebox{\venn}
\savebox{\venn} {\Huge
\begin{tikzpicture}
  \def\aonly{114}
  \def\bonly{169}
  \def\ab{60}
  \pgfmathsetmacro\a{\aonly + \ab}
  \pgfmathsetmacro\b{\bonly + \ab}

  \pgfmathsetmacro\ra{sqrt(\a)}
  \pgfmathsetmacro\rb{sqrt(\b)}

  \pgfmathsetmacro\scale{0.13}
  \pgfmathsetmacro\R{(\ra)*\scale}
  \pgfmathsetmacro\S{(\rb)*\scale}

  \pgfmathsetmacro\d{1.9} 

  \draw[fill=red!20, opacity=0.5] (0,0) circle (\R);
  \draw[fill=blue!20, opacity=0.5] (\d,0) circle (\S);

  \node[align=center] (Alabel) at (-\R -4, 0) {$\langle$all-MiniLM-L7-v2, (350, 8),\\ similarity, xlm-roberta-large$\rangle$};
  \node[align=center] (Blabel) at (\S+5, 0) {$\langle$LaBSE, (560, 5),\\similarity, tinyroberta$\rangle$};

  \node at (-0.3, 0) {\aonly};       
  \node at (\d+0.3, 0) {\bonly};     
  \node at (\d/2, 0) {\ab};          
\end{tikzpicture}
}

\begin{figure}[t]
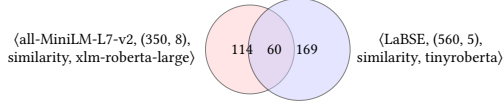

    \centering
    \resizebox{0.8\linewidth}{!}{
    {\usebox{\venn}}}
    \vspace{-0.1in}
    \caption{Failure modes of \lcqa configurations are uncorrelated.}
    \vspace{-0.2in}
    \label{fig:venn}
\end{figure}

Further, the estimated confidence score is a reliable indicator of the likelihood of the configuration being correct, which is exploited by \toolname to switch to a different configuration when needed.  To illustrate, here is a question and the corresponding answers provided by 4 of the configurations, in increasing order of confidence.  The incorrect answers have been crossed out.

\begin{itemize}[leftmargin=*, label=Q:]
  \item ``Human Error'' is the season finale of the third season of a TV show that aired on what network? 
\end{itemize}
\begin{enumerate}[leftmargin=*, label=$A_{\arabic*}$:]
    \item \sout{USA Network}. $\langle$all-MiniLM-L6-v2, (1400, 2), mmr, xlm-roberta-large$\rangle$ with confidence $0.00$. 
  \item \sout{House}. $\langle$all-MiniLM-L6-v2, (350, 8), mmr, xlm-roberta-large$\rangle$ with confidence $0.05$. 
  \item \sout{ABC}. $\langle$all-MiniLM-L6-v2, (350, 8), similarity, xlm-roberta-large$\rangle$ with confidence $0.30$.
  \item Fox. $\langle$LaBSE, (700, 4), mmr, xlm-roberta-large$\rangle$ with confidence $0.76$.
\end{enumerate}
The configuration corresponding to $A_3$ is actually the single best configuration overall (recall that it achieves an $F_1$ of $0.39$).  Yet, here it has low confidence in its answer, which also happens to be incorrect.  Further, it is not the QA model that needs to be changed, but the embedding model and the VectorStore (use larger chunks)---see configuration corresponding to $A_4$.

Contrary to intuition, it is not always the larger models that provide the correct answer or have the most confidence.
For the question ``The Oberoi family is part of a hotel company that has a head office in what city?'' the heavier model in $\langle$all-MiniLM-L6-v2, (350, 8), similarity, xlm-roberta-large$\rangle$ does extract the correct answer (``Delhi'') but with a confidence of only 0.01.  So \toolname turns to the lighter model in $\langle$all-MiniLM-L6-v2, (350, 8), similarity, tinyroberta$\rangle$, which gives the same answer but with a confidence of $0.98$.
Similarly for \icp,
Figure~\ref{fig:effective}(b)
shows that the different \icp configurations can reach a maximum $F_1$
metric of 0.90, but when using operator choices with \toolname, the $F_1$
metric reaches 0.94. The reason for this improvement is again uncorrelated
failures between the configurations. We skip more qualitative analysis due
to space constraints.

\textit{Overall, we conclude that using operator choices using \toolname could
improve performance by up to 23\%.}

\vspace{-5pt}
\subsection{Runtime Performance of \toolname}
\label{sec:runtime}

Since existing systems lack native support for operator choices, assertions, and
rollbacks, the accuracy advantage shown above could be achieved only by running
the workflow configurations sequentially. 
We create this setup by writing Spark and Flink jobs for each of the 20
configurations in the reduced \icp workflow described above. All the cards
are input to the workflow configuration with the highest $F_1$ score. The
outputs are checked with the same assertion as in Figure~\ref{fig:apps}(d).
Cards that fail the assertion are input to the next configuration. This
iterative process is continued until all the configurations are exhausted,
effectively simulating rollback. \toolname was evaluated using the same input
cards to compare the total processing time. Each system is limited to 10 CPU
cores. Each experiment was repeated 5 times. The variation in runtimes was
negligible.



\begin{figure}[t]
    \includegraphics[width=\linewidth]{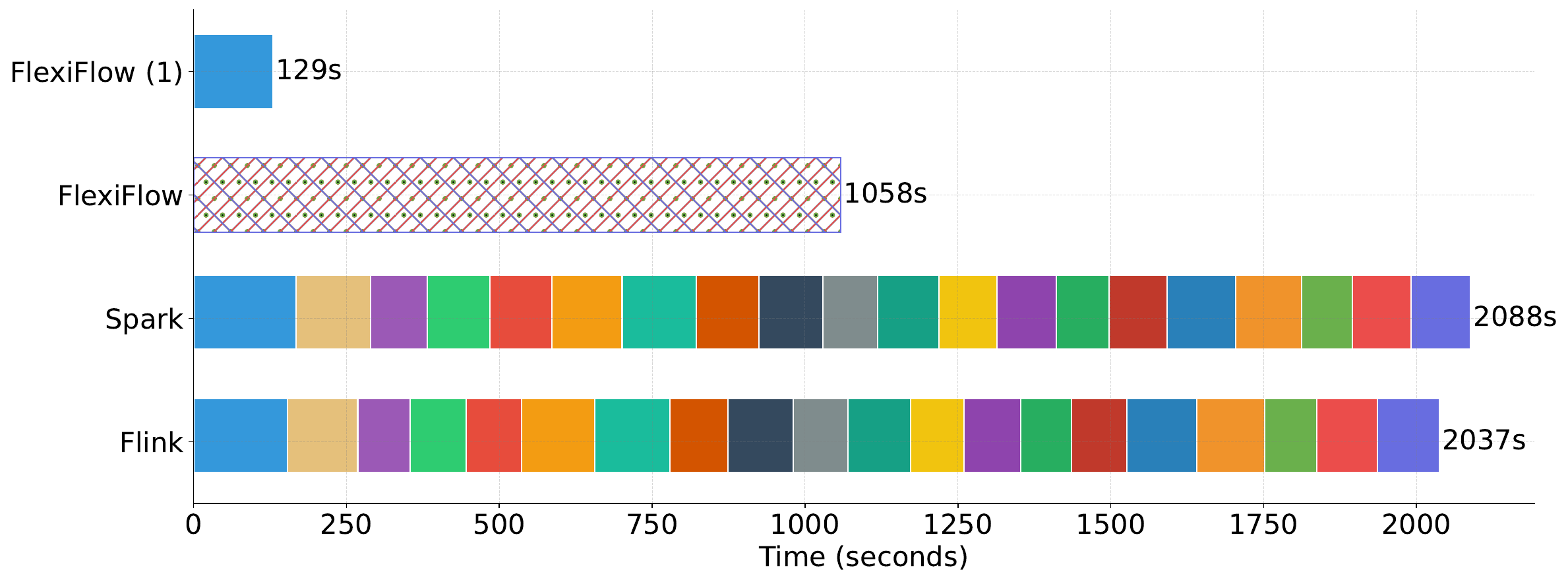}
    \vspace{-0.2in}
  \caption{Comparing runtime of \toolname with simulating rollback in other dataflow systems.}
    \label{fig:vs_sequential}
    \vspace{-0.2in}
\end{figure}


Figure~\ref{fig:vs_sequential} compares the total runtime for \toolname with
sequential execution of workflow configurations in Spark and Flink. Runtime of
each execution in the sequence is shown in a different color for Spark and
Flink. The figure shows that \toolname completes the run 48\% faster than
sequential executions in Flink and 49\% faster than those in Spark. The 
reason for this improvement is that \toolname can reuse intermediate results
during rollbacks and can better leverage data parallelism as it can run
different configurations for different inputs simultaneously.



We also compare the general performance of \toolname with Spark and Flink by
running just the best \icp configuration while disabling all the operator
choices and assertions from \toolname on identical inputs and setup.
Figure~\ref{fig:vs_sequential} shows this as \toolname(1). \toolname can finish
this run 15\% faster than Flink and 22\% faster than Spark\footnote{ This
performance difference might be because our implementations are written in
Python. This may cause Python-JVM interaction overheads for Spark and Flink.}.

\textit{Overall, we conclude that the general performance of \toolname is
comparable to state-of-the-art systems. The native support for operator choices,
assertions, and rollbacks in \toolname can improve runtimes by 48\% compared
to sequentially executing workflow configurations.}

\vspace{-0.1in}
\section{Discussion}
\label{sec:discussion}

\toolname reduces programmer burden by letting them specify operator choices
instead of maintaining multiple workflow instances. \toolname
allows a diversity of choices. 
\toolname reduces management burden of prioritizing
between multiple workflow instances using bandit-based strategies. 


This paper only takes the first step in letting programmers rely on \toolname
strategies to pick the right choice at each operator. In future work, the
proposed strategies can also take input data characteristics into account to
potentially learn different optimal sequences of configurations for data with
different characteristics (\eg cards from different countries in \icp).
\toolname is most effective when failure modes of configurations are
uncorrelated. But, failures may not be completely uncorrelated. For example, when
a heavy model fails to predict correctly, it might be more likely that the light
model will also fail. \toolname strategies do not separately learn
conditional probability of passing assertions which may be crucial for such
cases. Modeling conditional probabilities without managing large amounts of
state is also part of our future work. \toolname strategies currently focus on
lowering runtimes. But many batch workflows are more dollar-cost sensitive than
they are runtime sensitive, such as call commercial LLMs less often.
We believe a similar problem formulation may suffice for reducing dollar costs.
\vspace{-0.1in}
\section{Related Work}
\label{sec:relatedwork}

\vspace{5pt}
\noindent {\bf Workflow and DataFlow Systems.} 
\toolname is a dataflow
system where the operator order within DAG defines the data dependencies. The
operators follow pipeline execution in that operators execute as soon as their
input data is available. This is similar to Dataflow systems such as
MapReduce~\cite{dean2004mapreduce} and others~\cite{isard2007dryad,
zhou2012scope}. However these systems do not have support for alternative
operator choices, assertions, and rollbacks to improve on overall workflow
accuracy. Timely dataflow~\cite{murray2013naiad} allows for some form of rollback, but without alternative operator choices. 

The rollback and incremental computation in \toolname is similar to incremental updates
in incremental view maintenance~\cite{ivm1986, ivm1995, dbtoaster_icdm_2016, dbtoaster_vldb_2012}, streaming
computation~\cite{google_dataflow_vldb_2015, dbsp2023vldb}, and provenance-based selective
replay~\cite{provenance_predicates_2011}. These systems typically handle updates
at the input, such as base table updates in incremental view maintenance or
stream appends in streaming computation. 
In contrast, \toolname allows updates at intermediate outputs during rollbacks.

\vspace{5pt}
\noindent {\bf Programming Models.}
Operator choices are similar to McCarthy's ambiguous functions
(Amb)~\cite{mccarthy1959basis}  where the language runtime tries to pick an
alternative to satisfy assertions. Operator choices are also similar to choice
blocks in template programs~\cite{sketch2006asplos, sygus2013fmcad} where
programmers just write a template and let an SMT solver choose the right block
so that the overall program meets a specification. 
To the best of our knowledge, \toolname is the first to apply the concept
of operator choices within the context of a dataflow system. 
The target use case of \toolname is different from previous work.

\vspace{5pt}
\noindent {\bf DataFlow Debugging.} Systems such as Amber~\cite{kumar2020amber} and those
based on Spark including Titian~\cite{provenance_titian_2015},
BigSift~\cite{gulzar2018bigsift}, BigDebug~\cite{gulzar2016bigdebug}, and
TagSniff~\cite{contreras2019tagsniff} support debugging primitives such as
breakpoints (to pause and resume execution) and watchpoints (to inspect
intermediate data). While these can be used for error detection, often they
require manual changes for error correction. Similarly, model
assertions perform runtime checks to detect model errors at
runtime~\cite{kang2020model}.
However, model assertions work in the context of a single model
whereas we detect errors on overall workflows and correct them by rolling back
to alternate workflow configurations. 

\vspace{5pt}
\noindent \textbf{Online Learning.} 
\toolname strategies build upon
Thompson Sampling~\cite{Thompson1933, Sutton1998, russo2020tutorialthompsonsampling}, which samples all actions that could plausibly be optimal,
while shifting sampling away from those that are unlikely to be optimal.
Theoretically, TS has been shown to achieve asymptotically optimal logarithmic
regret in stochastic MAB settings~\cite{russo2020tutorialthompsonsampling}.
While a formal regret analysis for our approach is beyond the scope of the
current paper, experimentally we benefit from learning the right choices. We
also note that, while there are situations that do not require exploitation or
exploration, or situations where it is better to exploit a high performing
suboptimal action than to invest resources exploring actions, often modeling
such situations depends upon availability of some high-quality historical data.
We currently do not assume access to any such historical data and we learn in an online setting. 

\vspace{5pt}
\noindent \textbf{Confidence Estimation.} When the assertion in \toolname is a confidence threshold (as in \ic, \scqa, \lcqa), our system is related to the work done in confidence estimation which is applied to, e.g., out-of-distribution detection and selective prediction.  Examples of such work include Bayesian neural networks, ensemble of neural networks, and temperature scaling (see, e.g., the surveys \cite{gawlikowski2023survey, geng2023survey}.  Unlike our work, these do not attempt to mitigate the runtime cost of the ensemble by learning an optimal sequence of models to try. 
\vspace{-0.25in}
\section{Conclusion}
\label{sec:conclusion}

We showed that in many ML pipelines, relying on a single model to perform inference across all data batches often fails to maximize accuracy and thus overall performance. Our \toolname system, incorporates operator choices, assertions, and rollbacks and uses a novel multi-armed bandit approach to perform automatic model switching and thus improves overall accuracy and efficiency of 
various ML pipelines such as document processing,
image classification, and question-answering tasks.

Moving forward, our approach opens new avenues for AutoML-based systems that autonomously adapt to evolving data distributions. Future work will explore 
incorporating intermediate data characteristics, such as the country of an ID card,
into the learning strategies. Ultimately, \toolname represents a step toward more
intelligent, data-aware workflow systems that reduce manual intervention and
enhance the reliability of ML-driven automation.

\bibliographystyle{ACM-Reference-Format}
\begin{CJK*}{UTF8}{gbsn}
\bibliography{references}
\end{CJK*}

\appendix


\end{document}